\documentclass{article}

\usepackage{authblk}
\usepackage{microtype}
\usepackage{graphicx}
\usepackage{caption}
\usepackage{subcaption}
\usepackage{booktabs} %
\usepackage{hyperref}
\usepackage{natbib}
 \newlength\titlebox 
\usepackage{url}

\usepackage{amssymb}
\usepackage{epstopdf}
\usepackage{amsmath}
\usepackage{amsthm}
\usepackage{bm}

\usepackage[ISO]{diffcoeff}
\usepackage{optidef}
\usepackage{xspace}
\usepackage{mathtools}
\renewcommand{\cite}{\citep}
\newtheorem{thm}{Theorem}
\newtheorem{prop}[thm]{Proposition}
\newtheorem{lemma}[thm]{Lemma}
\newtheorem{cor}[thm]{Corollary}
\newtheorem{deff}[thm]{Definition}

\newtheorem{ex}[thm]{Example}
\newtheorem{assum}[thm]{Assumption}

\newcommand{\hf}{\hat{f}}
\newcommand{\hG}{\hat{G}}

\newcommand{\bv}{\mathbf{v}}
\newcommand{\cA}{\mathcal{A}}

\newcommand{\cD}{\mathcal{D}}
\newcommand{\cF}{\mathcal{F}}

\newcommand{\cJ}{\mathcal{J}}

\newcommand{\cM}{\mathcal{M}}

\newcommand{\cS}{\mathcal{S}}

\newcommand{\cP}{\mathcal{P}}

\newcommand{\cV}{\mathcal{V}}

\newcommand{\cX}{\mathcal{X}}

\newcommand{\bbR}{\mathbb{R}}
\newcommand{\bbN}{\mathbb{N}}
\newcommand{\bbZ}{\mathbb{Z}}
\newcommand{\bbE}{\mathbb{E}}

\newcommand{\ie}{\textit{i.e.}}
\newcommand{\eg}{\textit{e.g.}}

\newcommand{\incomplete}{\textcolor{red}{(INCOMPLETE)}}

\newcommand{\todo}[1]{\textcolor{red}{(TODO) #1}}

\newcommand{\train}{\mathrm{train}}
\newcommand{\test}{\mathrm{test}}

\newcommand{\piJ}{J_{\mathrm{PI}}}
\newcommand{\isJ}{J_{\mathrm{IS}}}
\newcommand{\drJ}{J_{\mathrm{DR}}}
\newcommand{\hpi}{{\hat{\pi}}}

\DeclareMathOperator*{\argmax}{argmax}
\DeclareMathOperator*{\argmin}{argmin}
\newcommand{\displayparen}[3]{\left #1 #2 \right #3}
\newcommand{\rbr}[1]{( #1 )}
\newcommand{\cbr}[1]{\{ #1 \}}
\newcommand{\sbr}[1]{[ #1 ]}

\newcommand{\abs}[1]{| #1 |}
\newcommand{\norm}[1]{\| #1 \|}
\newcommand{\rbrdisplay}[1]{\displayparen({#1})}
\newcommand{\cbrdisplay}[1]{\displayparen\{{#1}\}}
\newcommand{\sbrdisplay}[1]{\displayparen[{#1}]}

\newcommand{\normdisplay}[1]{\displayparen\|{#1}\|}
\newcommand{\rmd}{\mathrm{d}}

\newcommand{\bfm}{\mathbf{m}}
\newcommand{\bfv}{\mathbf{v}}
\newcommand{\bfu}{\mathbf{u}}

\date{}
\begin{document}

\title{Biases in \emph{In Silico} Evaluation of Molecular Optimization Methods and \\Bias-Reduced Evaluation Methodology}
\author[1]{Hiroshi Kajino, Kohei Miyaguchi, and Takayuki Osogami}
\affil[1]{IBM Research - Tokyo}
\maketitle
\begin{abstract}
 We are interested in \emph{in silico} evaluation methodology for molecular optimization methods.
 Given a sample of molecules and their properties of our interest, we wish not only to train an agent that can find molecules optimized with respect to the target property but also to evaluate its performance.
 A common practice is to train a predictor of the target property on the sample and use it for both training and evaluating the agent.
 We show that this evaluator potentially suffers from two biases; one is due to misspecification of the predictor and the other to reusing the same sample for training and evaluation.
 We discuss bias reduction methods for each of the biases comprehensively, and empirically investigate their effectiveness.
\end{abstract}
\section{Introduction}

Molecular optimization aims to discover novel molecules with improved properties,
which is often formulated as a reinforcement learning problem by modeling the construction of a molecule using a Markov decision process. %
The performance of such agents is measured by the quality of generated molecules.
In the community of machine learning, most of the molecular optimization methods have been verified \emph{in silico}, \ie, in computer simulation.
Since most of the generated molecules are novel, their properties are unknown and we have to resort to a predictor to estimate the properties.
However, little attention has been paid to how reliable such estimates are, which makes the existing performance estimates less reliable.
In this paper, we study the statistical performance of such performance estimators to enhance our understanding of the evaluation protocol and we discuss several directions to improve it.

Let us first introduce a common practice to estimate the performance \emph{in silico}.
Let $\cS^\star$ be a set of molecules, $f^\star\colon\cS^\star\to\bbR$ be a function evaluating the target property of the input molecule, and $\cD=\{(m_n,f^\star(m_n))\in\cS^\star\times\bbR\}_{n=1}^N$ be a sample.
A common practice is to train a predictor $f(m;\cD)$ using $\cD$, regard it as the true property function, and follow the standard evaluation protocol of online reinforcement learning.
That is, an agent is trained so as to optimize the properties of discovered molecules computed by $f(m;\cD)$, and its performance is estimated by letting it generate novel molecules and estimating their properties by $f(m;\cD)$.
We call this a \emph{plug-in performance estimator}~(Section~\ref{sec:plug-perf-estim}).

Our research question is \emph{how accurate the plug-in performance estimator is as compared to the true performance computed by $f^\star$}.
We first point out that the plug-in performance estimator is biased in two ways, indicating that it is not reliable in general~(Section~\ref{sec:two-biases-plug}).
The first bias called a \emph{model misspecification bias} comes from the deviation between the predictor and the true property function evaluated over the molecules discovered by the learned agent.
This bias is closely related to the one encountered in covariate shift~\cite{SHIMODAIRA2000227}.
It grows if molecules discovered by the agent become dissimilar to those used to train the predictor.
The second bias called a \emph{reusing bias} is caused by reusing the same dataset for training and testing the agent.
This bias has been pointed out in various areas~\cite{Ormoneit2002,pmlr-v80-ito18a}.
Due to these biases, the plug-in performance estimator is not necessarily a good estimator of the true performance.

We then discuss how to reduce these two biases.
Section~\ref{sec:reduc-missp-bias} introduces three approaches to reducing the misspecification bias.
Since the misspecification bias is caused by covariate shift, it can be reduced by training the predictor taking it into account~(Section~\ref{sec:covariate-shift}) and/or by constraining the agent so that the generated molecules become similar to those in the sample~(Section~\ref{sec:pessimism}).
Yet another approach is to use a more sophisticated estimator called a doubly-robust performance estimator~(Section~\ref{sec:doubly-robust-perf-1}).

Our idea to correct the reusing bias comes from the analogy to model selection~\cite{10.5555/1554702}, whose main objective is to estimate the test performance by correcting the bias of the training performance, \ie, the performance computed by reusing the same dataset for training and testing.
Given the analogy, one may consider train-test split allows us to reduce the reusing bias.
We however argue that it is not as effective as that applied to model selection due to the key difference between our setting and model selection; the test set in model selection is used to take expectation, while that in our setting is used to train a predictor, which is much more complex than expectation.
This complexity introduces a non-negligible bias to the train-test split estimator, resulting in a less accurate bias estimation~(Section~\ref{sec:bias-estim-train}).
We instead propose to use a bootstrap method in Section~\ref{sec:bootstr-bias-estim}, which is proven to estimate the reusing bias more accurately than the train-test split method.

We empirically validate our theory in Section~\ref{sec:empirical-studies}.
First, we quantify the two biases, and confirm that both of them are non-negligible, and the reusing bias increases as the sample size decreases, as predicted by our theory.
Second, we assess the effectiveness of the bias reduction methods, and confirm that the reusing bias can be corrected, while the reduction of the misspecification bias comes at the cost of performance degradation of the agent.

\paragraph{Notation.}
For any distribution $G$, let $\hG\sim G^N$ denote the empirical distribution of a sample of $N$ items independently drawn from $G$.
For a set $\cX$, let $\delta_x$ be Dirac's delta distribution at $x\in \cX$.
For any integer $M\in\bbN$, let $[M]\coloneqq\{0,\dots,M-1\}$.
For any set $A$, $\cP(A)$ denotes the set of probability distributions defined over $A$.

\section{Problem Setting}
We define a class of molecular optimization problems using a Markov decision process~(MDP) of length $H+1$~($H\in\bbN$).
Let $\cS$ be a set of states, and $s_\bot\in\cS$ be the terminal state.
Let $\cS^\star\subseteq\cS$ be a subset of states that correspond to valid molecules and the rest of the states correspond to incomplete representations of molecules.
Let $\cA$ be a set of actions that transform a possibly incomplete molecule into another one.
There exists the terminal action $a_\bot\in\cA$ that evaluates the property of the molecule at step $H$, after which the state transits to the terminal state $s_\bot$.
For $h\in[H+1]$, let $T_h\colon\cS\times\cA\rightarrow\cP(\cS)$ be a state transition distribution at step~$h$. %
Let $r_h\colon\cS\times\cA\rightarrow\bbR$ be a reward function at step $h$. %
Let $\rho_0\in\cP(\cS)$ be the initial state distribution.

We assume that the set of states at step $H$ is limited to $\cS^\star$,
and the reward function is defined as,
\begin{align*}
 r_h(s,a) = \begin{cases}
	     0 & h=0,1,\dots,H-1\\
	     f^\star(s) & h=H, a=a_\bot, s\in\cS^\star,
	    \end{cases}
\end{align*}
where we call $f^\star\colon\cS^\star\rightarrow\bbR$ a \emph{property function}, which measures the target property of the input molecule.
Let $\cM=\{\cS, \cA, \{T_h\}_{h=0}^H, \rho_0, H\}$ be the dynamical model of the MDP.
Throughout the paper, we assume we know $\cM$ and omit the dependency on it in expressions.

Let $\Pi$ be the set of policies and $\pi=\{\pi_h(a\mid s)\}_{h=0}^H\in\Pi$ be a policy modeled by a probability distribution over $\cA$ conditioned on $s\in\cS$.
At each step $h\in[H+1]$, the agent takes action $a_h$ sampled from $\pi_h(\cdot\mid s_h)$.
The performance of a policy is measured by the expected cumulative reward:
\begin{align*}
 J^\star(\pi)\coloneqq \bbE^\pi\left[\sum_{h=0}^H r_h(S_h,A_h)\right]=\bbE^\pi \left[f^\star(S_H)\right],
\end{align*}
where $\bbE^\pi[\cdot]$ is the expectation with respect to the Markov process induced by applying policy $\pi$ on $\cM$.
Letting $p_h^\pi(s)\in\cP(\cS^\star)$ be the distribution of states visited by policy $\pi$ at step $h\in[H+1]$, the expected cumulative reward is alternatively expressed as,
$ J^\star(\pi) = \bbE_{S\sim p_H^\pi} f^\star(S)$.

In practice, the property function is not available and instead a sample from it, $\cD=\{(m_n,f^\star(m_n))\in\cS^\star\times\bbR\}_{n=1}^N$, is available.
Let us assume that each tuple is independently distributed according to $G\in\cP(\cS^\star\times\bbR)$.
For a theoretical reason clarified in Section~\ref{sec:assumptions}, we use the empirical distribution of the sample, $\hG\in\cP(\cS^\star\times\bbR)$, rather than the sample itself~(Assumption~\ref{asm:smooth_RL_algo}) and we call $\hG$ an empirical distribution and a sample interchangeably.

Let us define a \emph{policy learner} $\pi\colon\cP(\cS^\star\times\bbR)\to\Pi$, an algorithm to learn a policy from a distribution over $\cS^\star\times\bbR$.
It typically receives a sample $\hG$ and outputs a policy, which we denote $\pi(\hG)$.
Our objective is to evaluate its performance by $J^\star(\pi(\hG))$ given access to $\pi$, $\hG$, and $\cM$.

We conclude this section by providing several examples that can or cannot be handled by our formulation.
Intuitively, a generative process that is guaranteed to generate a valid molecule within $H$ steps can be handled in our formulation.

\begin{ex}[String-based generation]
 Early attempts to generate molecules~\cite{bombarelli2016} represent a molecule by a string called SMILES~\cite{Weininger:1988aa}.
 Since the generated SMILES string is not necessarily valid, it cannot be handled within our framework.
 Another string representation called SELFIES~\cite{Krenn_2020} can be always decoded into a valid molecule; hence, a generative process using it can be handled by our formulation.
\end{ex}

\begin{ex}[Atom-wise generation]
 Another approach to generating a molecule is to start from void and add atoms one by one~\cite{li2018learning,you2018}.
 By prohibiting actions that lead to invalid molecules, we can guarantee that the generated molecules are always valid.
\end{ex}

\begin{ex}[Molecular generation by chemical synthesis]
 A sequential application of chemical reactions to a molecule can be modeled by our MDP, where each state corresponds to a molecule and each action corresponds to a possible reaction to it.
 In particular, \citet{pmlr-v119-gottipati20a} used a template-based chemical reaction, where an action consists of selecting a reaction template and selecting reactants.
\end{ex}

\section{Biases of Plug-in Performance Estimator}\label{sec:bias-plug-perf}
A widely used approach to estimating $J^\star(\pi(\hG))$ is a \emph{plug-in performance estimator}~(Section~\ref{sec:plug-perf-estim}).
We point out that it is biased in two ways~(Section~\ref{sec:two-biases-plug}) and theoretically characterize these biases in Sections~\ref{sec:missp-bias}~and~\ref{sec:reusing-bias}.

\subsection{Plug-in Performance Estimator}\label{sec:plug-perf-estim}
A \emph{plug-in performance estimator} is a common practice in the molecular optimization community to evaluate the performance of policy $\pi$ given a sample only.
Let us define,
\begin{align}
 \piJ(\pi,f) \coloneqq \bbE^\pi[f(S_H)],
\end{align}
where PI stands for \emph{plug-in}.
Let $f\colon\cP(\cS^\star\times\bbR)\to\bbR$ be an algorithm to train a predictor, typically by minimizing the loss function averaged over the input distribution.
Let $\hpi\coloneqq\pi(\hG)$ be a policy trained using $\hG$ and $\hf\coloneqq f(\hG)$ be a predictor trained using the same $\hG$.
Then, the plug-in performance estimator is defined as $\piJ(\hpi,\hf)$, which is often used as a proxy for the true performance, $J^\star(\hpi)$.

\subsection{Bias Decomposition}\label{sec:two-biases-plug}
The plug-in performance estimator is biased in two ways; the first bias comes from model misspecification of the predictor, and the second one is due to reusing the same sample for learning a policy and a predictor.
Let us define,
\begin{align*}
J(G_1,G_2) \coloneqq& \piJ(\pi(G_1),f(G_2)),\\
\Delta(G_1,G_2) \coloneqq& \piJ(\pi(G_1), f(G_2)) - J^\star(\pi(G_1)).
\end{align*}
The quantity $J(G_1,G_2)$ denotes the estimated performance of a policy trained with distribution $G_1$ evaluated by a predictor trained with $G_2$,
and $\Delta(G_1,G_2)$ denotes the deviation of the estimated performance from the ground truth.
Then, the bias we care is denoted by $\bbE_{\hG\sim G^N}\Delta(\hG,\hG)$, which is decomposed as shown in Proposition~\ref{prop:bias-decomposition}.

\begin{prop}
 \label{prop:bias-decomposition}
 The bias is decomposed into a \emph{reusing bias} and a \emph{misspecification bias} as follows:
\begin{align}
\nonumber  &\bbE_{\hG\sim G^N} \Delta(\hG,\hG)\\
\nonumber =& \bbE_{\hG\sim G^N} [J(\hG, \hG) - J(\hG, G) + J(\hG,G) - J^\star(\hpi)]\\
\label{eq:12} =& \underbrace{\bbE_{\hG\sim G^N} [J(\hG, \hG) - J(\hG, G)]}_{\text{Reusing bias}} + \underbrace{\bbE_{\hG\sim G^N} \Delta(\hG, G)}_{\text{Misspecification bias}}.
\end{align} 
\end{prop}

\subsection{Misspecification Bias}\label{sec:missp-bias}
The squared misspecification $\Delta(\hG,G)^2$ is upperbounded by Jensen's inequality as,
\begin{align}
\label{eq:26}\Delta(\hG,G)^2 = \left[\bbE^{\hpi}(f(S_H; G) - f^\star(S_H))\right]^2  \leq \bbE_{S\sim p_H^\hpi}(f(S; G) - f^\star(S))^2,
\end{align}
where $f(s;G)$ denotes the prediction for state $s$ by $f(G)$.
Assuming that $f(G)=\argmin_{f}\bbE_{S\sim G} (f(S) - f^\star(S))^2$ holds,
the bias increases if $f(G)$ fails to predict the properties of molecules generated by policy~$\hpi$,
which occurs when the predictor is misspecified~(\ie, $f(G)\neq f^\star$) and $p_H^{\hpi}(s)$ and $G(s)$ are largely deviated~(\ie, the discovered molecules are not similar to those in the sample).

\subsection{Reusing Bias}\label{sec:reusing-bias}
The former term of Eq.~\eqref{eq:12},
\begin{align}
 \label{eq:11} b_N(G)\coloneqq \bbE_{\hG\sim G^N} [J(\hG, \hG) - J(\hG, G)],
\end{align}
quantifies the bias caused by reusing the same finite sample for training and testing a policy, which we call a reusing bias\footnote{The reusing bias is caused by sample reuse as well as the finiteness of the sample, which is clear when the policy is independent from $\hG$; the reusing bias still exists in such a case if $f(\hG)\neq f(G)$.}.
We will theoretically investigate its properties to find that $b_N(G)=O(1/N)$~(Proposition~\ref{prop:bias}) and $b_N(G)\geq 0$~(Proposition~\ref{prop:opt-bias}).
Our approach to obtaining the former result is to expand $J(\hG,\hG)$ and $J(\hG,G)$ around $(G,G)$ assuming that $\hG$ is close to $G$~(\ie, $N$ is sufficiently large).
The latter result is obtained by following the standard proof technique~\cite{Ormoneit2002,pmlr-v80-ito18a}.

\subsubsection{Assumptions}\label{sec:assumptions}
This section introduces a set of technical assumptions for Proposition~\ref{prop:bias}.
Informally, we assume that (i) algorithms $\pi$ and $f$ depend on $\cD$ only through the empirical distribution $\hat{G}$ induced by sample $\cD$ and (ii) algorithms $\pi$ and $f$ are ``smooth'' with respect to the input distribution.
These are sufficient conditions to justify the expansion.
The first assumption is formally stated by Assumption~\ref{asm:smooth_RL_algo} with the notion of a normalized algorithm~(Definition~\ref{def:norm-algor}).

\begin{deff}[Normalized data-dependent algorithm]
 \label{def:norm-algor}
 A data-dependent algorithm $\alpha$, receiving data $\cD$ as input, is \emph{normalized} if its output $\alpha(\cD)$ depends on $\cD$ through its empirical distribution $\hat{G}$.
 If algorithm $\alpha$ is normalized, we denote its output by $\alpha(\hat{G})$.
\end{deff}

\begin{assum}
 \label{asm:smooth_RL_algo}
 Algorithms $f$ and $\pi$ are normalized. %
\end{assum}

The second assumption about the smoothness of the algorithms is formalized by \emph{entirety}.
An entire function\footnote{We regard any algorithm as a function.} allows Taylor-series expansion everywhere.

\begin{deff}[Entire function]
    \label{def:entire_function}
 Let $V$ and $W$ be Banach spaces.
 A function $\alpha\colon V\rightarrow W$ is \emph{entire}
 if there exist symmetric bounded multilinear maps
 $\{a_k:V^k\to W\}_{k=0}^\infty$ such that $\alpha(v)=\sum_{k=0}^\infty a_k(v^{\otimes k})~(v\in V)$
    and
    $\lim_{k\to \infty} \norm{a_k}^{1/k}=0$.
    Here,
    $\norm{a_k}\coloneqq \sup_{v_\ell\in V\setminus \cbr{0},1\le \ell\le k}\frac{a_k(v_1,...,v_k)}{\prod_{\ell=1}^k \norm{v_\ell}}$
    and $v^{\otimes k}$ denotes the $k$-repetition of $v$.
\end{deff}

\begin{assum}
 \label{assum:entire-algo} Algorithms $f$ and $\pi$ are entire.
\end{assum}

Finally, let us discuss that SGD-like algorithms satisfy the assumptions.
Informally, a normalized data-dependent algorithm that is defined by a sequence of noisy parameter updates is entire.
Since a family of SGD-based algorithms can be regarded as such algorithms, algorithms that train neural networks by SGD-like algorithms are regarded as entire.
See Appendix~\ref{sec:entirety-algorithms} for full discussion.

\subsubsection{Theoretical Characterizations}\label{sec:optimistic-bias}

Let us theoretically analyze the reusing bias, assuming the sample size $N$ is moderately large such that the asymptotic expansions are valid but $O(1/N)$ term cannot be ignored.
We show in Proposition~\ref{prop:bias} that the reusing bias is $O(1/N)$.
See Appendix~\ref{appendix:proofs-main-results} for its proof.

\begin{prop}%
 \label{prop:bias}
 Under Assumptions~\ref{asm:smooth_RL_algo} and \ref{assum:entire-algo},
 \begin{align*}
  b_N(G) = \frac1{2N}\bbE_{X\sim G} \left[2J^{(1,1)}_{G,G}(\delta_X-G,\delta_X-G) + J^{(0, 2)}_{G,G}(\delta_X-G, \delta_X-G)\right] +O\left(1/N^2\right) = O(1/N),
 \end{align*}
 holds  where $J_{G,G}^{(1,1)}$ and $J_{G,G}^{(0,2)}$ are the $(1,1)$-st and $(0, 2)$-nd Fr\'echet derivative of $J(G_1,G_2)$ at $(G_1,G_2)=(G,G)$.
\end{prop}

In particular, if the policy is optimal and the estimated property function is unbiased, \ie, $\bbE_{\hG\sim G^N} \hf = f(G)$~(which is true at least for a linear model),
we can prove that the bias is optimistic~(Proposition~\ref{prop:opt-bias}).
See Appendix~\ref{app:reusing-bias} for its proof.
\begin{prop}%
\label{prop:opt-bias}
 Assume $\bbE_{\hG\sim G^N} \hf = f(G)$ holds and 
 $\hpi=\argmax_{\pi\in\Pi}\piJ(\pi,\hf)$ holds.
 Then, $b_N(G)\geq 0$ holds.
\end{prop}

In summary, the sample reuse causes $O(1/N)$ bias, which is often optimistic.

\section{Bias Reduction Strategies}\label{sec:bias-reduct-strat}
We have witnessed that the plug-in performance estimator is biased in two ways. %
In this section, we discuss how to reduce these biases to obtain reliable performance estimates.

\subsection{Reducing Misspecification Bias}\label{sec:reduc-missp-bias}
There are mainly three approaches to reducing the misspecification bias, $\Delta(\hG,G)$.
The first one is to train the predictor considering the \emph{covariate shift}, a mismatch between training and testing distributions~(Section~\ref{sec:covariate-shift}).
The second approach is to constrain a policy such that the molecules discovered by the policy become similar to those in the sample $\hG$~(Section~\ref{sec:pessimism}).
These are mainly motivated by minimizing the right-hand side of Eq.~\eqref{eq:26}.
The third one is motivated by a standard technique in contextual bandit, the \emph{doubly-robust performance estimator} instead of the plug-in performance estimator~(Section~\ref{sec:doubly-robust-perf-1}).

Before going into details, let us introduce the notion of density ratio, which is used extensively to reduce the misspecification bias.
Let $G_0\in\Delta(\cS^\star)$ be any probability distribution over molecules whose support is larger than that of $p_H^\pi$.
Let $w^\star(s;\pi,G_0)={p_H^{\pi}(s)}/{G_0(s)}$ denote the density ratio between them, and let $w(s;\pi,G_0)$ denote a density ratio model approximating it.

\subsubsection{Covariate Shift}\label{sec:covariate-shift}
The misspecification bias can be reduced by minimizing the right-hand side of Eq.~\eqref{eq:26}, which is the mean squared error over $S\sim p_H^{\hpi}$.
The predictor $f(G)$ is usually trained by minimizing $\bbE_{S\sim G}(f(S;G) - f^\star(S))^2$ and does not necessarily minimize the right-hand side of Eq.~\eqref{eq:26} due to covariate shift~\cite{SHIMODAIRA2000227}, \ie, the mismatch between the training and testing distributions.
One approach suggested by the author to alleviating it is to train the predictor by weighted maximum-likelihood estimation.
Let $\ell_\lambda(f;\pi, w, G)=\bbE_{S\sim G} w(S;\pi, G)^\lambda (f(S) - f^\star(S))^2$ and 
\begin{align}
 \label{eq:19}f_\lambda(G_1,G_2) = \argmin_{f\in\cF} \ell_\lambda(f;\pi(G_1), w(G_2), G_2),
\end{align}
where $\lambda\in[0,1]$ controls the bias and variance of the estimated predictor\footnote{While $\lambda=1$ is optimal for $N\to\infty$, it will increase the variance for a finite sample size $N$, and a smaller $\lambda$ is favored.}.
By substituting $f_\lambda(G_1,G_2)$ for $f(G_2)$, the performance estimator is refined as,
\begin{align}
 \label{eq:17}J(G_1,G_2;\lambda) = \piJ(\pi(G_1), f_\lambda(G_1, G_2)),
\end{align}
which is expected to reduce the misspecification bias.

\subsubsection{Constrain a Policy}\label{sec:pessimism}
The first approach does not always work. If $p_H^\hpi$ and $G$ are not close enough, the effective sample size of the weighted maximum-likelihood estimation becomes small, leading to poor estimation.
This suggests that not all policy learners can be accurately evaluated; those whose state distribution $p_H^\pi$ is deviated from $G$ are difficult to be evaluated.

A straightforward idea to alleviate it is to constrain a divergence between $p_H^{\hpi}$ and $G$.
It is however computationally expensive, especially when $H$ is large.
We instead propose to regularize the policy by \emph{behavior cloning}~\cite{fujimoto2021a}.
Let us assume that the policy is obtained by solving the following optimization problem:
$
 \pi(G) = \argmin_{\pi\in\Pi} \ell(\pi;G),
$
and that there exists a behavior policy $\pi_0(\cdot\mid s)\in\cP(\cA)$ such that $p_H^{\pi_0}(s)=G(s)$ for $s\in\cS^\star$.
For $h\in[H+1]$, let $p_h^{\pi_0}(s,a)=p_h^{\pi_0}(s)\pi_0(a\mid s)$ and $p^{\pi_0}(s,a,h)=\frac{1}{H+1} p_h^{\pi_0}(s,a)$.
Then, our idea is to add the behavior cloning regularization as $\ell_\nu(\pi;G)\coloneqq \ell(\pi;G) - \nu \bbE_{(S,A,h)\sim p^{\pi_0}} {\log \pi_h(A\mid S)}$, and define
\begin{align}
\label{eq:20} \pi_\nu(G) \coloneqq \argmin_{\pi\in\Pi} \ell_\nu(\pi;G),
\end{align}
where $\nu\geq 0$ is a regularization hyperparameter.

Its empirical approximation can be computed without the behavior policy.
Assume that we can reconstruct a trajectory towards each molecule $s_H\in\cD$, namely $\tau=(s_0,a_0,s_1,\dots,s_H)$, and
let $\bar{\cD}$ denote the set of such trajectories.
Then, the empirical approximation of the regularization term is given by,
$\frac1{H|\bar{\cD}|}\sum_{(s_0,a_0,\dots,s_H)\in\bar{\cD}} \sum_{h\in[H]}\log \pi_h(a_h \mid s_h)$.
Although this regularization is not sufficient to constrain the divergence between $p_H^{\hpi}$ and $G$~(which has been discussed in the literature of imitation learning),
we consider behavior cloning is a simple yet effective heuristic, which will be investigated in the experiment.

\subsubsection{Doubly-Robust Performance Estimator}\label{sec:doubly-robust-perf-1}
The third approach to reducing the misspecification bias is a \emph{doubly-robust performance estimator}, which has been applied in contextual bandit~\cite{Dudik2014} and offline reinforcement learning~\cite{Tang*2020Doubly} as an alternative to the plug-in performance estimator.
Noticing that the performance can also be estimated via importance sampling, which we call an \emph{importance-sampling performance estimator},
the doubly-robust performance estimator combines these two estimators so as to inherit their benefits.

\paragraph{Importance-Sampling Performance Estimator.}\label{sec:import-sampl-perf}
Given the following change-of-measure,
\begin{align*}
 J^\star(\pi) &= \bbE^{\pi}f^\star(S_H) = \bbE_{S\sim G}\left(\frac{p_H^{\pi}(S)}{G(S)}\right)f^\star(S) = \bbE_{S\sim G}w^\star(S;\pi,G)f^\star(S),
\end{align*}
we obtain the importance-sampling performance estimator by substituting a density ratio model $w(s;\pi,G)$ for the true density ratio, resulting in,
\begin{align*}
 \isJ(\pi, w; G) \coloneqq \bbE_{S\sim G} w(S;\pi,G) f^\star(S).
\end{align*}
This coincides with $J^\star(\pi)$ if the density ratio model coincides with the true one.
By substituting the empirical distribution $\hG$ for $G$, we can compute it as follows:
\begin{align*}
 \isJ(\pi, w; \hG) = \bbE_{(S,f^\star(S))\sim\hG} w(S;\pi,\hG)f^\star(S).
\end{align*}

\paragraph{Doubly-Robust Performance Estimator.}\label{sec:doubly-robust-perf}
The doubly-robust performance estimator combines the plug-in and importance-sampling performance estimators as follows:
\begin{align}
 \label{eq:21}
\begin{split}
 \drJ(\pi, w, f;G) \coloneqq \bbE_{S\sim G} \left[w(S;\pi,G) (f^\star(S) - f(S))\right] + \bbE^{\pi} f(S_H). 
\end{split}
\end{align}
This estimator is a combination of the two estimators in that it is related to them as,
$\drJ(\pi, 0, f; G) = \piJ(\pi,f)$ and $\drJ(\pi, w, 0; G) = \isJ(\pi, w; G)$.

The misspecification bias is expressed as,
\begin{align*}
\begin{split}
 \Delta_{\mathrm{DR}}(\hG,G) \coloneqq \drJ(\hpi,w,f;G) - J^\star(\hpi) = \bbE_{S\sim G}(w(S;\hpi,G) - w^\star(S;\hpi,G))(f^\star(S)-f(S;G)), 
\end{split}
\end{align*}
which suggests that the misspecification bias disappears if the predictor or the density ratio model is correct.

\paragraph{Discussion.}
Notice that the misspecification biases of $\piJ$ and $\isJ$ are given by the followings:
\begin{align*}
 \Delta_{\mathrm{PI}}(\hG,G) \coloneqq& \piJ(\hpi,f) - J^\star(\hpi) = \bbE_{S\sim G}\left[w^\star(S;\hpi,G)(f(S;G) - f^\star(S))\right],\\
 \Delta_{\mathrm{IS}}(\hG,G) \coloneqq& \isJ(\hpi,w;G) - J^\star(\hpi) = \bbE_{S\sim G}\left[(w(S;\hpi,G) - w^\star(S;\hpi,G))f^\star(S)\right].
\end{align*}
We can deduce that for $S\sim p_H^\pi$ (i) if $|f^\star(S)-f(S;G)| \ll |f^\star(S)|$ holds, the misspecification bias of $\drJ$ will be smaller than that of $\isJ$,
and (ii) if $|w^\star(S;\hpi,G) - w(S;\hpi,G)|  \ll |w^\star(S;\hpi,G)|$ holds, the misspecification bias of $\drJ$ will be smaller than that of $\piJ$.
Therefore, if we can learn both of the predictor and the density ratio model well, the doubly-robust performance estimator is preferred to the other estimators.
Otherwise, the doubly-robust performance estimator can be worse than the others.

\subsubsection{Summary}
We have introduced three approaches to reducing misspecification bias.
The first one trains the predictor by weighted maximum likelihood estimation~(Eq.~\eqref{eq:19}).
The second one constrains the policy by behavior cloning~(Eq.~\eqref{eq:20}).
The third one is the doubly-robust performance estimator~(Eq.\eqref{eq:21}).
Taking these into consideration, let us re-define,
\begin{align}
\label{eq:14} J(G_1,G_2) \coloneqq \drJ(\pi_\nu(G_1),w(G_1,G_2),f_\lambda(G_1, G_2);G_2).
\end{align}
We call the density ratio model and the predictor an \emph{evaluator}.
Note that Proposition~\ref{prop:bias} holds for the revised performance estimator~(Eq.~\eqref{eq:14}) by further assuming that $w$ is normalized and entire. 
Proposition~\ref{prop:opt-bias} holds for the importance sampling performance estimator by further assuming the unbiasedness of the density ratio model,
but we have not found natural assumptions for the doubly-robust performance estimator. See Appendix~\ref{app:reusing-bias} for details.

\subsection{Reducing Reusing Bias}
Our approach to reducing the reusing bias $b_N(G)$~(Eq.~\eqref{eq:11}) is to estimate the reusing bias and substract it from the estimator.
Such a bias reduction has been extensively discussed in the literature of information criteria~\cite{10.5555/1554702}, which aim to estimate the test performance of a predictor in a supervised learning setting by correcting the bias of its training performance. There are mainly two approaches: train-test split method and bootstrap method.

\subsubsection{Bias Estimation by Train-test Split}\label{sec:bias-estim-train}
The first approach estimates the bias via train-test split of the sample.
The sample $\cD$ is randomly split into $\cD_{\train}$ and $\cD_{\test}$ such that $\cD_\train\cap\cD_\test=\emptyset$ and $\cD_\train\cup\cD_\test=\cD$.
Let $\hG_{\train}$ and $\hG_\test$ denote the corresponding empirical distributions.
The reusing bias is estimated by
$ b_{\text{split}}(\hG) = \bbE \left[J(\hG_\train, \hG_\train) - J(\hG_\train,\hG_\test)\right]$, 
where the expectation is with respect to the random split of $\hG$.

While this estimator seems to be reasonable, it is not recommended due to the bias of the bias estimator.
As demonstrated in Proposition~\ref{prop:train-test-split-bias}, the train-test split estimator has $O(1/N)$ bias, the same order as the bias $b_N(G)$ itself,
and therefore, we cannot distinguish between the bias and the bias of the bias.
Such a bias is due to the non-linearlity of $J(G_1,G_2)$ with respect to $G_2$, the distribution used for testing\footnote{The standard supervised learning scenario does not suffer from this bias because the performance estimator is linear with respect to the testing distribution.}.
See Appendix~\ref{appendix:proofs-main-results} for its proof and Appendix~\ref{appendix:train-test-split} for the comparison with supervised learning.

\begin{prop}%
 \label{prop:train-test-split-bias}
Suppose we randomly divide the sample such that $|\cD_\train|:|\cD_\test|=\lambda:(1-\lambda)$ for some $\lambda\in(0,1)$.
 Under Assumptions~\ref{asm:smooth_RL_algo}~and~\ref{assum:entire-algo}, $\bbE_{\hG\sim G^N} [b_{\mathrm{split}}(\hG)] = b_N(G) + O(1/N)$ holds.
\end{prop}

Note that direct estimation of test performance by $J(\hG_\train,\hG_\test)$ is not recommended similarly, unless the size of the test sample is sufficiently large.
See Appendix~\ref{appendix:train-test-split} for detailed discussion.

\subsubsection{Bootstrap Bias Estimation}\label{sec:bootstr-bias-estim}
An alternative approach to estimating the reusing bias~(Eq.~\eqref{eq:11}) is bootstrap~\cite{efron1994introduction}.
A bootstrap estimator of the reusing bias $b_N(G)$ is obtained by plugging $\hat{G}$ into $G$:
\begin{align}
 \label{eq:2}b_N(\hat{G}) = \bbE_{\hG^\star\sim \hat{G}^N}\left[ J(\hat{G}^\star, \hat{G}^\star) - J(\hat{G}^\star, \hat{G}) \right].
\end{align}

Let $\hG^{(m)}$~($m\in[M]$) be a bootstrap sample obtained by uniform-randomly sampling data points $N$ times from the original sample $\cD$ with replacement.
Then, a Monte-Carlo approximation of Eq.~\eqref{eq:2} is,
$\hat{b}_N(\hat{G}) = \frac{1}{M} \sum_{m=1}^M \left[ J(\hat{G}^{(m)}, \hat{G}^{(m)}) - J(\hat{G}^{(m)}, \hat{G}) \right]$.

In contrast to the train-test split method, the bootstrap bias estimation can estimate the bias as stated in Proposition~\ref{prop:bootstr-bias-estim}.
See Appendix~\ref{appendix:proofs-main-results} for its proof.
\begin{prop}%
 \label{prop:bootstr-bias-estim}
 Under Assumptions~\ref{asm:smooth_RL_algo}~and~\ref{assum:entire-algo}, $\bbE_{\hG\sim G^N} [b_N(\hG)] = b_N(G) + O(1/N^2)$ holds.
\end{prop}

\subsubsection{Summary}
We have introduced two reusing-bias estimators, referring to the literature of information criteria.
We have found that the train-test split estimator, one of the most popular estimators, cannot reliably estimate the bias in our problem setting, although it works in supervised learning.
In contrast, the bootstrap bias estimator is shown to be less biased than the train-test split estimator and can estimate the reusing bias more reliably.
Therefore, we conclude that the bootstrap bias estimator is preferable to the train-test split estimator.

From computational point of view, the bootstrap bias estimator requires us to train $M$ agents and $M+1$ evaluators.
We set $M=20$ in the experiments given the result of a preliminary experiment.
Since the bootstrap procedure can be easily parallelized with low overhead, its wall-clock time can be reduced in proportion to the computational resource.

\section{Empirical Studies}\label{sec:empirical-studies}
In this section, we aim to empirically quantify the two biases as well as the effectiveness of the bias reduction methods.

\subsection{Setup}
We describe our experimental setup.
See Appendix~\ref{appendix:exper-sett} for full details to ensure reproduciability.

\textbf{Molecular Representation.}
Unless otherwise indicated, all of the functions defined over molecules use the 1024-bit Morgan fingerprint~\cite{Morgan1965,Rogers2010} with radius 2 as a feature extractor. %

\textbf{Environment and Agent.}
We employ the environment and the agent by \citet{pmlr-v119-gottipati20a} with minor modifications.
The agent receives a molecule as the current state, and outputs an action consisting of a reaction template and a reactant.
The environment, receiving the action, applies the chemical reaction defined by the action to the current molecule to generate a product, which is then set as the next state. This procedure is repeated for $H$ times, and lastly the agent takes action $a_\bot$ to be rewarded by the property of the final product. We set $H=1$ to reduce the variance in the estimated performance and better highlight the biases and their reduction.
The agent is implemented by actor-critic using fully-connected neural networks. %

We use the reaction templates curated by \citet{Button2019} and prepare the reactants from the set of commercially available substances in the same way as the original environment.
The number of reaction templates is 64, 15 of which require one reactant, and 49 of which require two reactants. The number of reactants is 150,560.

\textbf{Evaluators.}
As a predictor, we use a fully-connected neural network with one hidden layer of 96 units  with softplus activations except for the last layer.
It is trained by minimizing the risk defined over $S\sim G$. %

As the density ratio model, we use the kernel unconstrained least-squares importance fitting~(KuLSI)~\cite{Kanamori2012}. %
In particular, we use the trained predictor except for the last linear transformation as a feature extractor and compute the linear kernel using it.

\textbf{Evaluation framework.}
To evaluate the biases, we need the true property function~$f^\star$, which however is not available in general.
We thus design a semi-synthetic experiment using a real-world dataset $\cD_0=\{(m_n,f^\star(m_n))\in\cS^\star\times\bbR\}_{n=1}^{N_0}$. %
We regard a predictor trained with $\cD_0$ as the true property function~$f^\star$.
In specific, we used the predictor provided by \citet{pmlr-v119-gottipati20a}, which was trained with the ChEMBL database~\cite{Gaulton2017} to predict $\mathrm{pIC}_{50}$ value associated with C-C chemokine receptor type 5~(CCR5).
With this property function, we have full access to the environment, and
we can construct an offline dataset~$\cD$ of an arbitrary sample size by running a random policy on $\cM$, which is available in our setting.

To decompose the bias into the misspecification bias and the reusing bias, we need $f(G)$, the predictor obtained with full access to the data-generating distribution $G$.
We approximate it by $f(\hG_\test)$, where $\hG_\test$ is the empirical distribution induced by a large sample $\cD_\test$ of size $10^5$ constructed independently of $\cD$.
This approximation is valid if $|\cD_\test|$ is sufficiently large~(see Proposition~\ref{prop:train-test-split-large-test}).
Then, the misspecification bias can be estimated by $J(\hG, \hG_\test) - J^\star(\hpi)$ and the reusing bias by $J(\hG,\hG) - J(\hG, \hG_\test)$.
The performance estimators are defined by the expectation with respect to a trajectory of a policy, and we estimate them by Monte-Carlo approximation with 1,000 trajectories.

\begin{figure*}[t]
\centering
\begin{subfigure}[t]{0.3\textwidth}
 \includegraphics[width=\hsize]{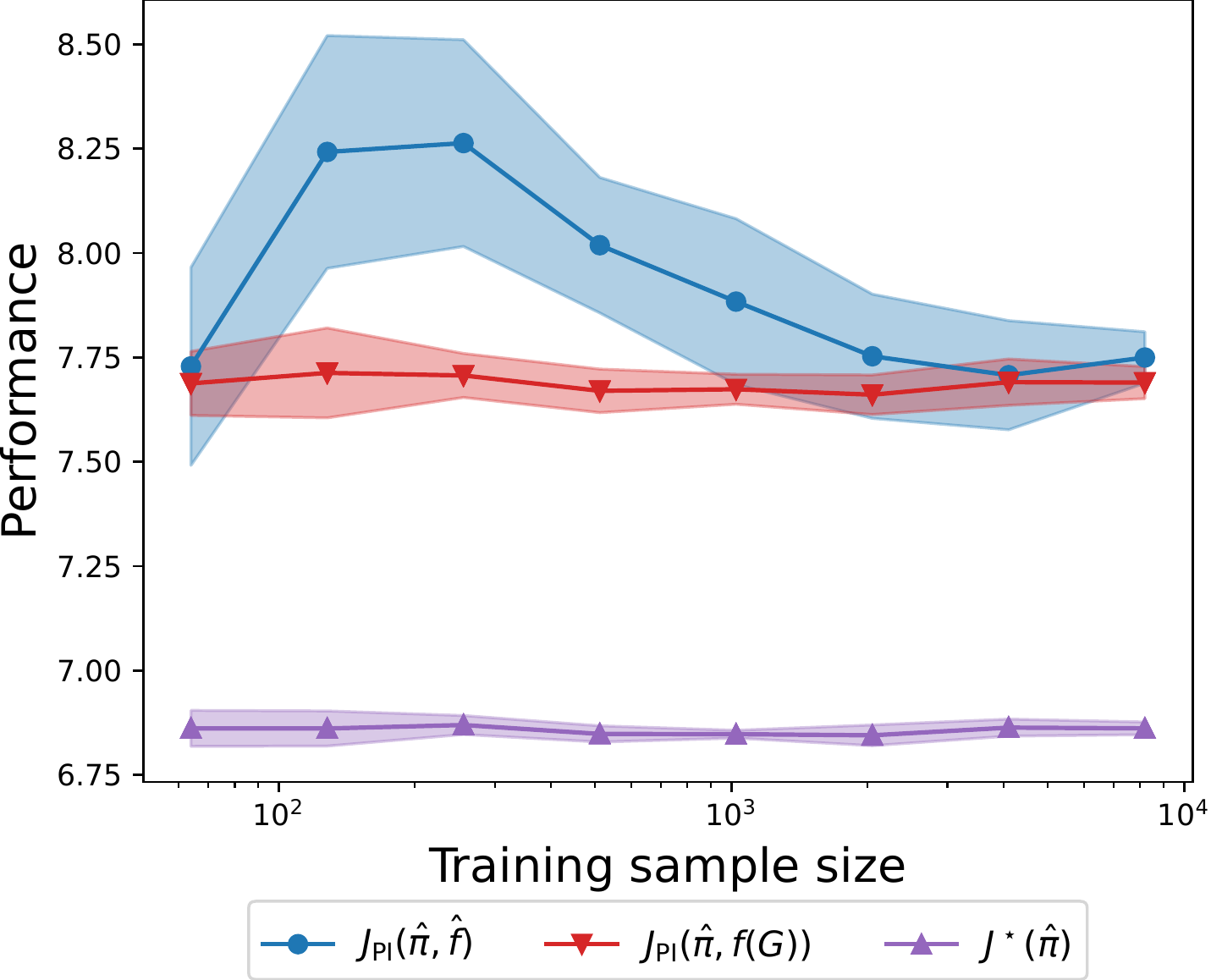}
\end{subfigure}
\hspace{0.03\textwidth}
 \begin{subfigure}[t]{0.3\textwidth}
\includegraphics[width=\hsize]{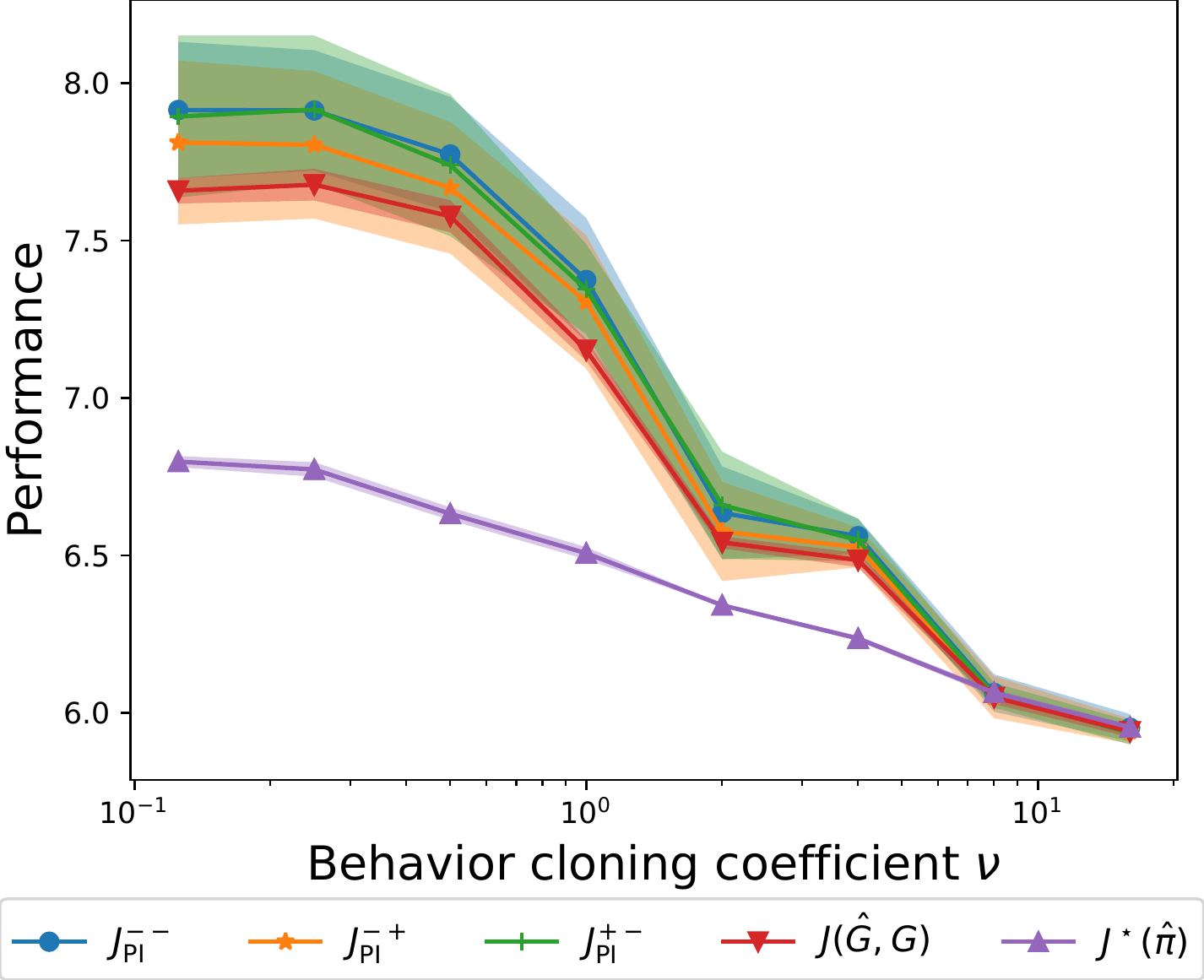}
 \end{subfigure}
 \hspace{0.03\textwidth}
 \begin{subfigure}[t]{0.3\textwidth}
 \includegraphics[width=\hsize]{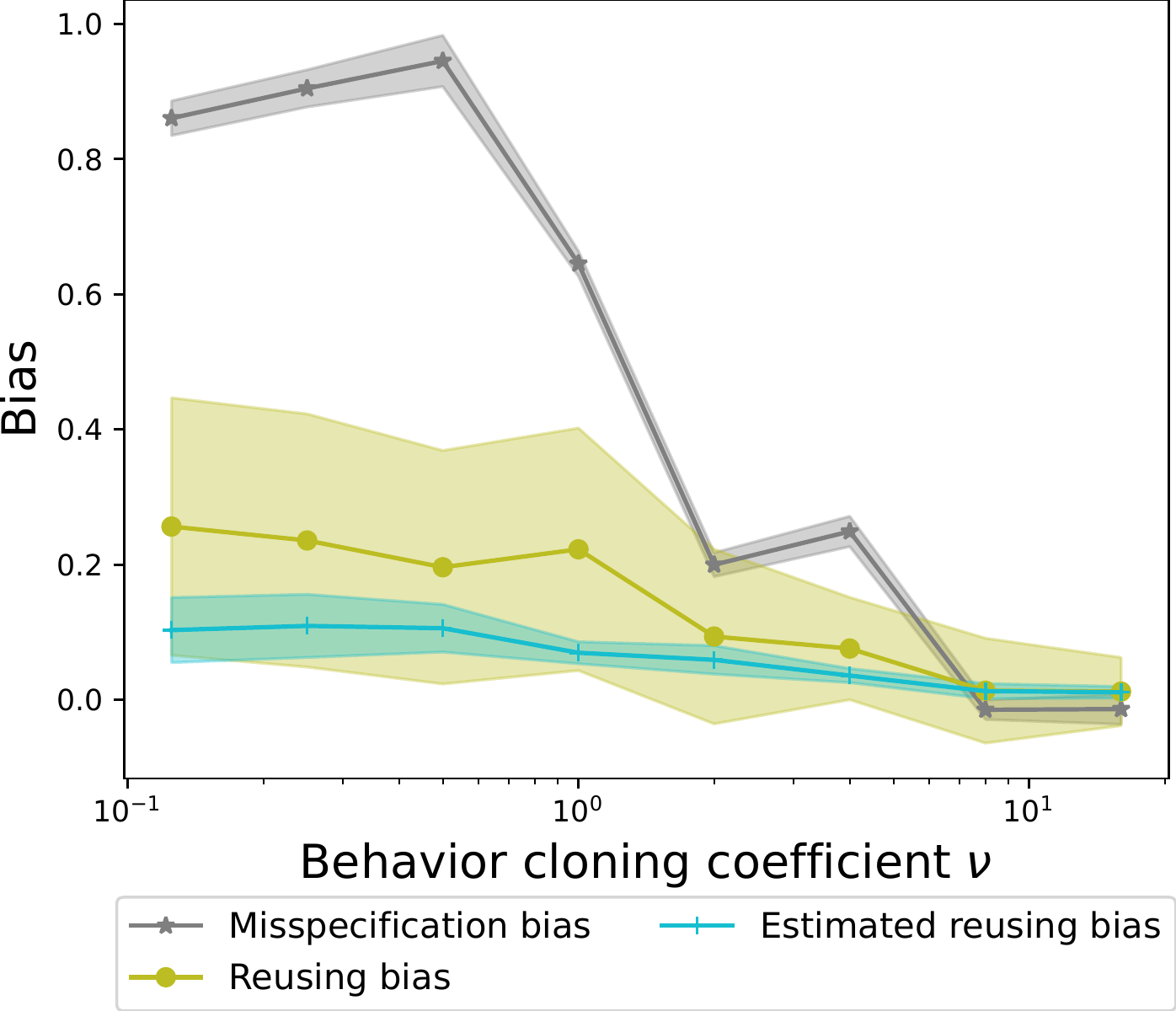}
 \end{subfigure}
\caption{Lines show means and shaded areas show standard deviations.
 \textbf{(Left)} Relationship between the biases and the sample size. $\piJ(\hpi,\hf)-\piJ(\hpi,f(G))$ corresponds to the reusing bias and $\piJ(\hpi,f(G)) - J^\star(\hpi)$ to the misspecification bias.
 \textbf{(Middle)} Comparison between bias reduction methods. 
\textbf{(Right)} Comparison between the misspecification bias, reusing bias, and the estimated reusing bias.
 }
 \label{fig:exp-res}
\end{figure*}

\subsection{Quantifying the Two Biases}
First, we aim to quantify the misspecification and reusing biases.
Assuming we fix the algorithms $\pi$, $f$, and $w$, the biases are mostly influenced by the sample size of the training set.
Thus, we design an experiment to investigate the relationship between these biases and the sample size.

We vary the training sample size $N$ in $\{2^6, 2^7,\dots,2^{13}\}$.
For each $N$, we repeatedly generate a pair of train and test sets for {five times}, and evaluate the biases as indicated above.
We report the means and standard deviations.

Figure~\ref{fig:exp-res}~(left) illustrates the result.
We have three observations.
First, when $N=2^7$, the misspecification bias, $\piJ(\hpi,f(G)) - J^\star(\hpi)$,  was roughly twice as large as the reusing bias, $\piJ(\hpi,\hf)-\piJ(\hpi,f(G))$, demonstrating that both are non-negligible.
Second, for $N\geq 2^7$, the reusing bias increased as the size of the training sample decreased, which coincides with Proposition~\ref{prop:bias}.
The results for $N < 2^7$ did not coincide with it because the sample size is not large enough for asymptotic expansion to be justified.
Third, the ground-truth performance of the policies were rather stable across different training sample sizes.
We found that the policies were similar to each other, suggesting that this environment has a local optimum with a reasonably good performance.
In fact, the performance of a random policy is around 5.8, and we have not found other better policies.
This also suggests that the policy learner was insensitive to the particular sample, and the reusing bias in this case is mainly caused by the finiteness of the sample to train the predictor, not by reusing the same sample.

\subsection{Quantifying Bias Reduction Methods}
We then study the effectiveness of the bias reduction methods presented in Section~\ref{sec:bias-reduct-strat}.
Since the behavior cloning coefficient $\nu$ will control the trade-off between the misspecification bias and the performance of the learned policy,
it should be determined according to the user's requirement, \ie, whether the accuracy of performance estimation or the actual performance is prioritized.
Therefore, we design an experiment to evaluate the effectiveness of the bias reduction methods, varying $\nu$ in the range of $\{2^{-4},\dots,2^4\}$.

Let $\piJ^{b_1 b_2}~(b_1, b_2\in\{+,-\})$ be the plug-in performance estimator with covariate shift~($b_1=+$) or without it ($b_1=-$) and with bootstrap bias reduction~($b_2=+$) or without it~($b_2=-$).
Let us define $\drJ^{b_1 b_2}$ accordingly for the doubly-robust performance estimator.
We compare the performance estimates by $\piJ^{--}$, $\piJ^{+-}$, $\piJ^{-+}$, and $\drJ^{--}$ to see the effectiveness of each of the bias reduction methods.

Figure~\ref{fig:exp-res}~(middle) illustrates the performance estimates for $N=10^3$.
Since $\drJ^{--}$ performs significantly worse than the baseline $\piJ^{--}$, we omit it from the figure. See Appendix~\ref{sec:addit-exper-result} for the full result.
We observe that the bootstrap bias reduction worked well, while the benefit of the covariate shift strategy is marginal.
This indicates that the density ratio estimation did not work well in this setting.

Figure~\ref{fig:exp-res}~(right) illustrates the biases in $\piJ^{--}$ and the reusing bias estimated by the bootstrap method.
As we expected, the misspecification bias tends to decrease as we increase $\nu$.
The reusing bias is under-estimated, but the estimated reusing bias contributes to bias correction.

In summary, we confirm that (i) behavior cloning can reduce the misspecification bias at the expense of performance degradation, (ii) the reusing bias can be estimated and corrected by bootstrap, and (iii) the methods using density ratio models did not perform well in our setting.

\section{Related Work}
Our primary contribution is the comprehensive study of theoretically-sound evaluation methodology for molecular optimization algorithms using real-world data.
Since the pioneering work by \citet{bombarelli2016},
a number of studies on this topic have been published in the communities of machine learning and cheminformatics to advance the state-of-the-art.
While some of them~\cite{bombarelli2016a,10.1145/3394486.3403346,Das2021} have been validated \emph{in vitro}, many others have been evaluated \emph{in silico}. %

Early studies~\cite{pmlr-v70-kusner17a} adopted the octanol-water partition coefficient, $\log P$, penalized by the synthetic accessibility score~\cite{Ertl2009} and the number of long rings as the target property to be maximized. The score can be easily computed by RDKit, and is often implicitly regarded as a reliable score computed by an accurate simulator.
Some recently consider that the $\log P$ optimization is not appropriate as a benchmark task because it is easy to optimize~\cite{doi:10.1021/acs.jcim.8b00839} or its prediction can be inaccurate~\cite{yang2021practical}, and alternative benchmark tasks have been investigated; some of them propose a suite of benchmark tasks~\cite{doi:10.1021/acs.jcim.8b00839,10.3389/fphar.2020.565644} and the others use other property functions trained by real-world data~\cite{Olivecrona2017,Li2018,pmlr-v119-jin20b,pmlr-v119-gottipati20a,xie2021mars}.

Our contribution to this line of studies is that we empirically and theoretically demonstrate potential biases in the current evaluation methodology using real-world data and present bias reduction methods.
This also unveils why the $\log P$ optimization task has been hacked and suggests that the alternative benchmark tasks will be hacked as long as no bias reduction method is applied.
The $\log P$ function implemented in RDKit~\cite{Wildman1999} is obtained by fitting a linear model to a dataset of experimental $\log P$ values, and is in fact a predictor.
Our theory suggests that  unless the bias reduction methods are applied, the learned agent generates unrealistic molecules that are far from those in the dataset~(which has been often reported in $\log P$ optimization), and the resultant performance estimates are biased.
It also suggests that by incorporating bias reduction methods, we can reliably estimate the performance and therefore can safely compare different methods even when using the $\log P$ optimization task.

Our work shares a similar objective with a seminal work by \citet{pmlr-v80-ito18a}, which aims to reduce the reusing bias that appears when solving an optimization problem whose parameters are estimated from data.
A major contribution to this literature is to relax their assummption that the predictor is correctly specified.
This introduces the concept of misspecification bias, which was confirmed to be non-negligible in our application.
Another minor contribution is to formalize their reusing-bias correction method by bootstrap and investigate the theoretical properties.

\section{Conclusion and Future Work}
We have discussed that the plug-in performance estimator is biased in two ways; one is due to model misspecification and the other is due to reusing the same dataset for training and testing.
In order to reduce these biases to obtain more accurate estimates, we recommend to (i)~add a constraint to the policy such that the state distribution stays close to the data distribution and (ii)~correct the bias by bootstrapping if it is non-negligible and we can afford to do it.

A future research direction is to improve the density ratio estimation so that the other bias reduction methods work.
Another direction is to constrain a policy with less performance degradation.
Since the methods using variational autoencoders~\cite{bombarelli2016,pmlr-v80-jin18a,pmlr-v97-kajino19a} can naturally generate molecules similar to those in the data,
such methods could be reevaluated.

\newpage
\bibliography{main}
\bibliographystyle{plainnat}

\newpage
\onecolumn

\appendix

\section{Technical Background}
For completeness, let us first define the Fr\'echet derivative and Taylor expansion using it in Appendix~\ref{appendix:tayl-expans-using}, and
let us discuss several basic properties of entire functions in Appendix~\ref{appendix:prop-entire-funct}.

\subsection{Taylor Expansion using Fr\'echet Derivative}\label{appendix:tayl-expans-using}
In this paper, we mainly analyze a function between Banach spaces by its Taylor expansion.
To do so, it is necessary to introduce the Fr\'echet derivative, which is a generalization of the total derivative on the space of real numbers to that on Banach spaces.
In this section, we provide a brief introduction to the Fr\'echet derivative and Taylor expansion.

Let $V$ and $W$ be Banach spaces, $U\subset V$, and $f\colon U\to W$ be a function.
If a bounded linear mapping $A\colon V\to W$ such that
\begin{align}
 \label{eq:4}\lim_{\|h\|\rightarrow 0} \frac{\|f(x+h) - f(x) - A_x(h)\|}{\|h\|} = 0
\end{align}
exists, $f_x^{(1)} \coloneqq A_x$ is called the Fr\'echet derivative of $f$ at $x\in U$.
Let $D$ be the Fr\'echet differential operator and we express $Df_x \coloneqq f_x^{(1)}$ when emphasizing the operator.
 Equation~\eqref{eq:4} implies that,
\begin{align}
 f(x+h) = f(x) + f_x^{(1)}(h) + o(\|h\|),
\end{align}
holds. Similarly, we can define a higher-order Fr\'echet derivative $f_x^{(k)}$ for $k\geq 0$, and it is a symmetric multilinear map from $V^k$ to $W$ when fixing $x$.
The Taylor expansion of $f$ is obtained as,
\begin{align}
 f(x+h) = \sum_{k=0}^\infty \frac1{k!} f_x^{(k)}(h^{\otimes k}),
\end{align}
where $h^{\otimes k}$ represents $k$ repetitions of $h$.

For a bivariate function $f(x,y)\colon U^2\to W$, let us introduce partial Fr\'echet derivatives.
If a bounded linear mapping $A\colon V\to W$ such that,
\begin{align}
 \lim_{\|h_x\|\rightarrow 0} \frac{\|f(x+h_x,y) - f(x,y) - A_{x,y}(h_x)\|}{\|h_x\|} = 0
\end{align}
exists, $f_{x,y}^{(1,0)}\coloneqq A_{x,y}$ is called the $(1,0)$-th Fr\'echet derivative of $f$ at $(x,y)\in U$ with respect to $x$.
Similarly, we can define the $(k,l)$-th Fr\'echet derivative $f_{x,y}^{(k,l)}$ as a multilinear map from $V^{k}\times V^l$ to $W$, when fixing $(x,y)\in U^2$.
Let $D^{(k,l)}$ be the Fr\'echet differential operator and we express $D^{(k,l)}f_{x,y}\coloneqq f_{x,y}^{(k,l)}$ when putting emphasis on the operator.
Then, the Taylor expansion of $f$ is obtained as,
\begin{align}
 f(x + h_x, y + h_y) = \sum_{k=0}^\infty \frac1{k!} \sum_{l=0}^k {k \choose l}f_{x,y}^{(l,k-l)}(h_x^{\otimes l},h_y^{\otimes (l-k)}).
\end{align}

\subsection{Properties of Entire Functions}\label{appendix:prop-entire-funct}
We present several properties of entire functions, which will be used to prove our main theoretical results.

The following proposition says that entire functions are closed under function composition.
\begin{prop}
    \label{prop:entirety_composition}
    Let $V$, $W$, and $U$ be Banach spaces.
    Let $f:V\to W$ and $g:W \to U$ be entire functions.
    Then, the composition $g \circ f$ is also entire.
\end{prop}
\begin{proof}
 Since $f$ and $g$ are entire, $f$ and $g$ can be expanded with coefficients $\{a_k\}_{k=0}^\infty$ and $\{b_k\}_{k=0}^\infty$ that decay super-exponentially.
    Define $h:V\to U$ as the formal expansion of $g\circ f$,
    \begin{align*}
        h(v)
        &\coloneqq
        \sum_{k=0}^\infty b_k\rbrdisplay{\rbrdisplay{\sum_{\ell=0}^\infty a_\ell(v^{\otimes \ell})}^{\otimes k}}
        \\
        &=
        \sum_{k=0}^\infty \sum_{\ell_1,...,\ell_k=0}^\infty b_k\rbrdisplay{\bigotimes_{i=1}^k a_{\ell_i}(v^{\otimes \ell_i})}
        \\
        &=
        \sum_{m=0}^\infty c_m(v^{\otimes m}),
    \end{align*}
    where $c_m:V^m\to U$ is the symmetric bounded multilinear map given by
    \begin{align*}
        c_m(v_1,...,v_m)
        &\coloneqq
        \sum_{k=0}^\infty \sum_{\ell_1+...+\ell_k=m} b_k\rbrdisplay{\bigotimes_{i=1}^k a_{\ell_i}(v_{\ell^{k-1}+1},...,v_{\ell^{k}})}, \quad m\ge 0,
    \end{align*}
    and $\ell^k\coloneqq \sum_{s=1}^k \ell_s$.
    Then, it suffices to show $h$ is entire, i.e., $\norm{c_m}^{1/m}\to 0$ as $m\to \infty$.
    By the assumption of entirety, there exist $C_a,C_b<\infty$ for all $r_a,r_b>0$
    such that
    \begin{align*}
        \norm{a_k}&\le C_a r_a^k,
                  &
        \norm{b_k}&\le C_b r_b^k
    \end{align*}
    for all $k\ge 0$.
    Thus, by the triangle inequality,
    \begin{align*}
        \norm{c_m}
        &\le
        \sum_{k=0}^\infty \norm{b_k} \sum_{\ell_1+...+\ell_k=m}\prod_{i=1}^k \norm{a_{\ell_i}}
        \\
        &\le
        \sum_{k=0}^\infty C_b r_b^k \sum_{\ell_1+...+\ell_k=m} C_a^k r_a^m
        \\
        &=
        C_b r_a^m\sum_{k=0}^\infty \sum_{\ell_1+...+\ell_k=m} (C_a r_b)^k
        \\
        &=
        C_b r_a^m\cbrdisplay{1+\sum_{k=0}^\infty {k+m \choose m} (C_a r_b)^{k+1}}.
        &(\text{$k\leftarrow k-1$})
    \end{align*}
    Let $S_m(r)\coloneqq \sum_{k=0}^\infty {k+m \choose m} r^k$
    and observe $S_m(r)=\frac{1}{(1-r)^{m+1}}$ for $0< r< 1$ according to
    the recursion starting with $S_0(r)=\frac{1}{1-r}$ and then
    $(1-r)S_m(r) = \sum_{k=0}^\infty {k+m-1 \choose m-1} r^k=S_{m-1}(r)$.
    Therefore, taking $r_b$ such that $C_a r_b < 1$,
    we have
    \begin{align*}
        \norm{c_m}
        &\le
        C_b\, r_a^m\sbrdisplay{1+C_a\, r_b\, S_m(C_a r_b)}
        \\
        &=
        C_b r_a^m\sbrdisplay{
            1+
            \frac{C_a r_b}{\rbr{1-C_a r_b}^{m+1}}
        },
        \quad m\ge 0.
    \end{align*}
    Since $r_a>0$ can be taken arbitrarily small, we have $\norm{c_m}^{1/m}\to 0$.
\end{proof}

\begin{prop}
    \label{prop:gevrey_condition}
    Let $f:V\to W$ be an infinitely Fr\'echet-differentiable map
    from a Banach space to another.
    Then $f$ is entire if and only if
    \begin{align*}
        \lim_{k\to \infty}\sup_{v\in K} \rbrdisplay{\frac{\norm{D^k f(v)}}{k!}}^{1/k} =0
    \end{align*}
    for all compact subset $K\subset V$. Here, $D$ denotes the Fr\'echet differential operator.
\end{prop}
\begin{proof}
    Let $a_k(v)\coloneqq D^kf(v)/k!$ be the $k$-th order coefficient of the Taylor expansion of $f$ at $v\in V$,
    $k\ge 0$.
    Also let $\sigma_k(v)\coloneqq \norm{a_k(v)}^{1/k}$.
    The goal of the proof is to show the equivalence between
    the entirety of $f$ and the uniform decay of $\sigma_k(v)$ on any compact $K\subset V$.

    First, we show the uniform decay implies the entirety.
    Take an arbitrary $v\in V$ with $v\neq 0$ and let $K_v\coloneqq \cbr{\lambda v:\lambda\in[0, 1]}$.
    Since $K_v$ is compact, we have $\lim_{k\to \infty} \sup_{u\in K_v} \sigma_k(u)= 0$,
    which implies there exists $C<\infty$ such that $\norm{a_k(u)}\le C(2\norm{v})^{-k}$ for all $u\in K_v$ and $k\ge0$.
    Now let $f_k$ be the $k$-th order finite partial sum of the Taylor expansion of $f$ at the origin,
    \begin{align*}
        f_k(u)&\coloneqq\sum_{\ell =0}^{k} a_\ell(0)(u^{\otimes \ell}), \quad k\ge 0, \quad u\in V.
    \end{align*}
    Then, there exists $u_0\in K_v$ for all $k\ge 0$ such that
    \begin{align*}
        \norm{f(v) - f_k(v)}
        &=
        \norm{a_{k+1}(u_0)(v^{\otimes k})}
        &(\because \text{Taylor's theorem})
        \\
        &\le
        \norm{a_{k+1}(u_0)}\norm{v}^k
        \\
        &\le
        \frac{C\norm{v}^k}{2^{k+1}\norm{v}^{k+1}}
        \overset{k\to \infty}{\longrightarrow} 0.
    \end{align*}
    Since the above argument holds for any $v\in V$ with $v\neq 0$ and trivially $f_k(0)=f(0)$,
    $f_k$ converges to $f$ everywhere.
    This implies the entirety of $f$ on $V$.

    Second, we show the entirety implies
    the uniform decay.
    For all $v,u\in V$, we have
    \begin{align*}
        f(v+u)
        &=
        \sum_{k=0}^\infty a_k(0)((v+u)^{\otimes k})
        \\
        &=
        \sum_{k=0}^\infty \sum_{\ell=0}^k {k\choose \ell } a_k(0)(v^{\otimes k-\ell}, u^{\otimes \ell})
        \\
        &=
        \sum_{\ell=0}^\infty \sum_{k=\ell}^\infty  {k\choose \ell } a_k(0)(v^{\otimes k-\ell}, u^{\otimes \ell})
        \\
        &=
        \sum_{\ell=0}^\infty a_\ell(v)(u^{\otimes \ell}).
    \end{align*}
    This implies, for all $v\in K$,
    \begin{align*}
        \norm{a_\ell(v)}
        &=
        \normdisplay{\sum_{k=\ell}^\infty  {k\choose \ell } a_k(0)(v^{\otimes k-\ell}, \cdot)}
        \le 
        \sum_{k=\ell}^\infty  {k\choose \ell } \norm{a_k(0)}R^{k-\ell},
    \end{align*}
    where $R\coloneqq \sup_{v\in K} \norm{v}$,
    which is finite since $K$ is compact.
    Note that the entirety of $f$ ensures the existence of $C<\infty$ for all $r>0$ such that $a_k(0)\le Cr^k$, $k\ge 0$.
    Take such a pair $(C, r)$ with $r<1/R$ and
    we have $\norm{a_\ell(v)}\le C r^\ell \sum_{k=0}^\infty {\ell + k\choose \ell}(rR)^k=\frac{C r^\ell}{(1-rR)^{\ell+1}}$,
    which implies $\limsup_{k\to \infty}\sup_{v\in K}\sigma_k(v)\le  r/(1-rR)$.
    Since $r>0$ can be taken arbitrarily small,
    the uniform convergence of $\sigma_k(v)$ to zero follows.
\end{proof}

\subsection{Entirety of Algorithms}
\label{sec:entirety-algorithms}
This section discusses entirety of algorithms.
First, let us prove that the plug-in performance estimator is entire with respect to input policy $\pi$ and evaluator $f$.

\begin{lemma}
    \label{lem:analyticity_of_cumulative_reward}
 $\piJ\colon\Pi\times\cF\to\bbR$ is entire, where $\Pi$ is a set of policies and $\cF\subseteq(\cS^\star\to\bbR)$ is a set of predictors.
\end{lemma}
\begin{proof}
 For any $\pi\in\Pi$ and $f\in\cF$, we have,
 \begin{align*}
  \piJ(\pi, f)=\bbE[ f(S_H) ]=\mathbf{p} P_0^\pi P_1^\pi \cdots P_H^\pi \mathbf{f},
 \end{align*}
    where
    $\mathbf{p}\coloneqq(\rho_0(s))_{s\in\cS}\in \bbR^{|\cS|}$
    is the row vector representing the initial state distribution,
    $P_h^\pi\coloneqq (\sum_{a\in \cA} T_h(s^\prime \mid s,a) \pi(a|s))_{s,s'\in\cS}\in \bbR^{|\cS|\times |\cS|}$
    is the transition matrix,
    and 
    $\mathbf{f}\coloneqq (f(s))_{s\in\cS}\in \bbR^{|\cS|}$ is the column vector representing the reward at the final step.
    Then, it is obvious that
    $\piJ(\pi, f)$ is a polynomial function, which is entire.
\end{proof}

Then, the following proposition shows that a sequential noisy linear computation defined by Equation~\eqref{eq:24} is entire.

\begin{prop}[Entirety of sequential noisy linear computation]
    \label{lem:stochastic_composition_lemma}
    Let $T$ and $d$ be positive integers.
    Let $\bfv=(v_1,...,v_T)\in V^T$
    be a sequence of input variables in a Banach space $V$,
    and 
    $\mathbf{E}=(\epsilon_1,...,\epsilon_T)\in \bbR^{d\times T}$
    be a sequence of $d$-dimensional noise vectors
    whose elements are drawn independently from the standard normal distribution.
    Let $\cbr{f_t:V\times \bbR^d\to \bbR^d}_{t=1}^T$ be a sequence of maps
    that are linear in the first argument and satisfy
    $\norm{f_t}\coloneqq \sup_{\norm{v}=1, \theta\in \bbR^d} \abs{f_t(v, \theta)}<\infty$,
    which is used to define the sequential computation
    \begin{align}
\label{eq:24}
\begin{split}
        \theta_0&\coloneqq 0,
        \\
        \theta_t&\coloneqq f_t(v_t, \theta_{t-1}) + \epsilon_t, \quad 1\le t\le T. 
\end{split}
    \end{align}
    Then,
    letting
    $g: \bbR^d\to W$ be an arbitrary bounded map to a Banach space $W$,
    the following function $h:V^T\to W$ is entire,
    \begin{align*}
        h(\bfv)\coloneqq \bbE_{\mathbf{E}}\sbr{g(\theta_T)}.
    \end{align*}
\end{prop}

Proposition~\ref{lem:stochastic_composition_lemma} implies that SGD-based algorithms can be considered to be entire.
For example, let us consider an SGD-based algorithm to minimize $\ell(\theta;\hG) = \bbE_{Z\sim \hG} \ell(\theta,Z)$, which operates as follows for $t=1,\dots,T$, starting from $\theta_0=0$:
\begin{align}
 \begin{split}
  \hG_t &\sim \hG^B,\\
  \theta_t &= \theta_{t-1} - \alpha_{t} \diffp{}{\theta} \bbE_{Z\sim \hG_t} \ell(\theta_{t-1},Z),
 \end{split}
\end{align}
where $\hG_t$ denotes a mini-batch of size $B$.
Theoretical studies of SGD algorithms~\cite{pmlr-v119-wu20c} often regard the noisy mini-batch gradient as the true gradient corrupted by a Gaussian noise, which is also supported by empirical studies~(for example, \citet{panigrahi2019nongaussianity} concluded that the gradient noise follows a Gaussian at least in the early stage of learning when the batch size is large).
Given this approximation, a mini-batch SGD algorithm is desribed as follows:
\begin{align}
 \label{eq:25}
 \begin{split}
  \theta_0 &= 0,\\
  \theta_t &= \theta_{t-1} - \alpha_{t} \diffp{}{\theta} \bbE_{Z\sim \hG} \ell(\theta_{t-1},Z) + \epsilon_t.
 \end{split}
\end{align}
Equation~\eqref{eq:25} can be handled by Proposition~\ref{lem:stochastic_composition_lemma} by setting $v_t=\hG$ and $f_t(\hG,\theta_{t-1})=\theta_{t-1} - \alpha_t \diffp{}{\theta}\bbE_{Z\sim \hG}\ell(\theta_{t-1}, Z)$, which is linear in the first argument and is bounded.
By setting $g$ as a function that maps from a parameter of the policy to a collection of $|\cS|$ probability measures in $\cP(\cA)$~(which is bounded for any neural network architecture), the whole algorithm $\pi(G)$ is described by the sequential noisy linear computation.

\begin{proof}
 [Proof of Proposition~\ref{lem:stochastic_composition_lemma}]
    Let $D_t$, $1\le t\le T$, denote the Fr\'echet differential operator on $v_t$ and
    $D^\bfm\coloneqq \prod_{t=1}^T D_t^{m_t}$ denote the multivariate higher-order counterpart,
    where $\bfm\coloneqq (m_1,...,m_T)\in \bbZ_{\ge 0}^T$ is multi-index.
    Define $c_m(\bfv)$ as the $m$-th order coefficient of the Taylor expansion of $h$ around $\bfv$,
    \begin{align}
        c_m(\bfv)(\bfu)&
        \coloneqq \sum_{\bfm:|\bfm|=m}\frac{D^{\bfm}h(\bfv)(u_1^{\otimes m_1},...,u_T^{\otimes m_T})}{\bfm!},\quad \bfu\in V^T,
        \label{eq:proof_coefficient_of_h}
    \end{align}
    where $|\bfm|\coloneqq \sum_{t=1}^T m_t$ and $\bfm!\coloneqq \prod_{t=1}^T m_t!$.
    Below, we show the entirety of $h$ through the decay rate of $c_m(\bfv)$.

    Define the functions $\cbr{g_t:\bbR^d\to W}_{t=1}^T$ representing the computation from step $t$ to $T$,
    starting with $g_T(\theta)\coloneqq g(\theta)$
    and then backward-recursively $g_{t-1}(\theta)\coloneqq g_t(f_{t-1}(v_{t-1}, \theta) + \epsilon_t)$ for $1\le t\le T$ and $\theta\in \bbR^d$.
    Then, we have the following decomposition formulae
    \begin{align*}
        g(\theta_T)\equiv g_t(\phi_t+\epsilon_t), \quad 1\le t\le T,
    \end{align*}
    where $\phi_t\coloneqq f_t(v_t, \theta_{t-1})$.
    Note that $g_t(\cdot)$, $\phi_t$ and $\epsilon_t$ are mutually statistically independent.
    Thus,
    letting $p(\epsilon)\coloneqq \frac1{\sqrt{2\pi}}e^{-\frac12\norm{\epsilon}_2^2}$, $\epsilon\in \bbR^d$ be the density function of $\epsilon_{t}$,
    we have
    \begin{align*}
        D_t h(\bfv)
        &= D_t \bbE\sbrdisplay{g(\theta_T)}
        \\
        &= D_t \bbE\sbrdisplay{g_t(\phi_t+\epsilon_t)}
        \\
        &= D_t \bbE\sbrdisplay{\int g_t(\phi_t+\epsilon)p(\epsilon)\rmd \epsilon}
        &(\because\text{independence among $g_t(\cdot), \epsilon_t, \phi_t$})
        \\
        &= D_t \bbE\sbrdisplay{\int g_t(\epsilon)p(\epsilon-\phi_t)\rmd \epsilon}
        \\
        &= \bbE\sbrdisplay{D_t f_t(v_t, \theta_{t-1})^\top \cdot \frac{\partial \int g_t(\epsilon)p(\epsilon-\phi_t)\rmd \epsilon}{\partial \phi_t} }
        &(\because \text{chain rule})
        \\
        &= \bbE\sbrdisplay{\xi_t(\theta_{t-1})^\top\int (\epsilon-\phi_t)g_t(\epsilon)p(\epsilon-\phi_t)\rmd \epsilon}
        \\
        &= \bbE\sbrdisplay{\xi_t(\theta_{t-1})^\top \int \epsilon g_t(\phi_t+\epsilon)p(\epsilon) \rmd \epsilon}
        \\
        &=\bbE\sbrdisplay{
            g(\theta_T)
            \epsilon_t^\top\xi_t(\theta_{t-1})
        },
    \end{align*}
    where
    $\xi_t(\theta)\coloneqq D_t f_t(v_t, \theta)$,
    which is constant with respect to $v_t$ since $f_t$ is linear on its first argument.
    Iterating the same procedure, we have
    \begin{align*}
        D_t^m h(\bfv)
        &=\bbE\sbrdisplay{
            g(\theta_T)\,
            \hat H_m(\epsilon_t;\xi_t(\theta_{t-1}))
        },\quad m\ge 0,
    \end{align*}
    where
    $\hat H_m(\epsilon;a):V^m\to \bbR$ is given by the recursion,
    starting with $\hat H_0(\epsilon;a)\coloneqq 1$,
    \begin{align*}
        \hat H_m(\epsilon;a)(u_1,...,u_m)
        &\coloneqq
        a(u_m)^\top \rbr{\epsilon-\nabla_\epsilon}\, \hat H_{m-1}(\epsilon;a)(u_1,...,u_{m-1}),
        \quad m\ge 1,
    \end{align*}
    for all
    $\epsilon\in\bbR^d$,
    $a:V\to\bbR^d$ and
    $u_1,...,u_m\in V$.
    Here, $\nabla_\epsilon$ denotes the gradient operator with respect to $\epsilon$.
    Further applying this procedure to multiple $t$s,
    we have
    \begin{align*}
        D^\bfm h(\bfv)
        &=\bbE\sbrdisplay{
            g(\theta_T)\,
            \prod_{t=1}^T \hat H_{m_t}(\epsilon_t;\xi_t(\theta_{t-1}))
        },\quad m\ge 0.
    \end{align*}
    Observe that, by induction with respect to $m\ge 0$,
    we have $\hat H_m(\epsilon;a)(u^{\otimes m})=\norm{a(u)}^mH_m(\epsilon^\top \widehat{a(u)})$, $u\in V$,
    where $\cbr{H_m(x)}_{m=1}^\infty$ is the (probabilist's) Hermite polynomials and $\widehat{x}=x/\norm{x}_2$ denotes the normalization of a vector $x\in\bbR^d$.
    Thus, for all $m\ge 0$,
    \begin{align}
        D^{\bfm}h(\bfv)(u_1^{\otimes m_1},...,u_T^{\otimes m_T})
        &=\bbE\sbrdisplay{
            g(\theta_T)\,
            \prod_{t=1}^T \norm{f_t(u_t,\theta_{t-1})}^{m_t}\, H_{m_t}(\epsilon_t^\top \widehat{f_t}(u_t, \theta_{t-1}))
        }.
    \end{align}
    Since the scaled Hermite polynomials $H_n(x)/\sqrt{n!}$ form an orthonormal basis of $L^2(p)$,
    by H\"older's inequality,
    \begin{align*}
        \abs{D^{\bfm}h(\bfv)(u_1^{\otimes m_1},...,u_T^{\otimes m_T})}
        &\le\norm{g}_{\infty}\prod_{t=1}^T\norm{f_t}^{m_t}\norm{u_t}^{m_t} \sqrt{m_t!},
    \end{align*}
    where $\norm{g}_{\infty}$ denotes the supremum norm of $g$.
    Substituting this back to Equation~\eqref{eq:proof_coefficient_of_h},
    we get an upper bound on $\norm{c_m(\bfv)}$,
    \begin{align*}
        \norm{c_m(\bfv)}
        &=\sup_{\bfu:\forall t, \norm{u_t}\le 1}\abs{c_m(\bfv)(\bfu)}
        \\
        &\le
        \norm{g}_{\infty}\sum_{\bfm:|\bfm|=m}\prod_{t=1}^T \frac{\norm{f_t}^{m_t}}{\sqrt{m_t!}}
        \\
        &\le
        \norm{g}_{\infty} {m+T-1\choose m}\frac{\max_{1\le t\le T}\norm{f_t}^m}{\sqrt{\Gamma(\frac{m}{T}+1)}},
        &(\because\text{convexity of $\ln \Gamma(x)$}),
    \end{align*}
    where $\Gamma(x)$ denotes the gamma function.
    This proves the super-exponential decay of $\norm{c_m(\bfv)}$ independent of $\bfv\in V^T$,
    which yields the lemma by Proposition~\ref{prop:gevrey_condition}.
\end{proof}

\subsection{Stochastic Expansion}
Stochastic expansion is a mathematical tool to expand an estimator~$\theta(G)$ with respect to its input distribution~$G$.
We are often interested in the estimator averaged over the possible sample space, $\bbE_{\hG\sim G^N}\theta(\hG)$,
and we often expand $\theta(\hG)$ around $G$ to understand the averaged estimator.
This section provides a useful formula to compute it.

\begin{lemma}[Stochastic expansion formula]
\label{lem:stochastic-exp-formula}
 Let $\cX$ be a set and let $\cP(\cX)$ be the set of probability measures on $\cX$.
 Let $V$ be a Banach space, and let $f\colon \cP(\cX)\to V$ be a function.
 Let $\cD=\{X_n\}_{n=1}^N$ be an i.i.d. sample from $G\in\cP(\cX)$. Let $n_1,\dots,n_k\in[N]$.
 If the $k$-th Fr\'echet derivative of $f$ exists at $G$ and there exists $i\in[k]$ such that $n_i\neq n_j$ for all $j\in[k]\backslash\{i\}$~(such an index $i$ is called \emph{singular}), then
\begin{align}
 \bbE_{\cD}[f_G^{(k)}(\delta_{X_{n_1}}-G,\dots,\delta_{X_{n_k}}-G)]=0,
\end{align}
 holds. Moreover, the number of assignments $(n_1,\dots,n_k)\in [N]^k$ such that there \emph{does not} exist singular indices is $O(N^{\lfloor k/2 \rfloor})$ regarding $k$ as a constant.
\end{lemma}

\begin{proof}
The expectation can be calculated as,
 \begin{align*}
  &\bbE_{\cD}[f_G^{(k)}(\delta_{X_{n_1}}-G,\dots,\delta_{X_{n_k}}-G)] \\
  = &\bbE_{\cD\backslash{X_i}}[\bbE_{X_i}[f_G^{(k)}(\delta_{X_{n_1}}-G,\dots,\delta_{X_{n_i}}-G,\dots,\delta_{X_{n_k}}-G)]] & (\because \{X_n\}_{n=1}^N \text{are independent}) \\
  = &\bbE_{\cD\backslash{X_i}}[f_G^{(k)}(\delta_{X_{n_1}}-G,\dots,\bbE_{X_i}[\delta_{X_{n_i}}-G],\dots,\delta_{X_{n_k}}-G)] & (\because f_G^{(k)} \text{is multilinear}) \\
  = &\bbE_{\cD\backslash{X_i}}[f_G^{(k)}(\delta_{X_{n_1}}-G,\dots,0,\dots,\delta_{X_{n_k}}-G)]  \\
  = & 0. & (\because f_G^{(k)} \text{is multilinear})
 \end{align*}

 \begin{figure}[t]
\centering
  \includegraphics[width=.8\hsize]{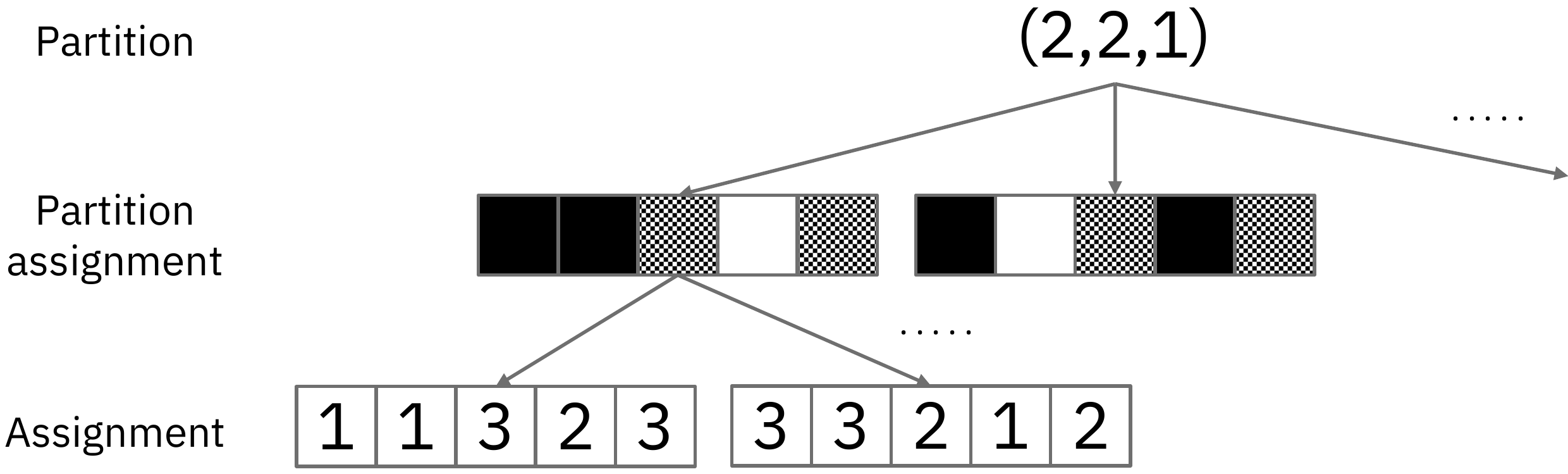}
\caption{Assignment in $N=3$, $k=5$ can be abstracted into a \emph{partition assignment}, where blocks with the same pattern fill will have the same index.
A partition assignment is further abstracted into a \emph{partition}, a sequence of non-increasing integers, each of which indicates the size of each pattern.}
\label{fig:assignment}
 \end{figure}
 
Then, let us count the number of non-singular assignments $(n_1,\dots,n_k)\in[N]^k$ by using the abstraction illustrated in Figure~\ref{fig:assignment}.
Assignments can be abstracted into \emph{partition assignments}, and they are further abstracted into \emph{partitions}, as explained in the caption of Figure~\ref{fig:assignment}.
 
A sequence of integers $p=(p_1,\dots,p_L)$ is a partition if and only if $p_1\geq\cdots\geq p_L\geq 1$ and $\sum_{l=1}^L p_l = k$.
The number of partitions depends only on $k$, not on $N$.
 For each partition, the number of associated partition assignments depends only on $k$, not on $N$.
 For each partition assignment with partition $p=(p_1,\dots,p_L)$, the number of associated assignments is at most $N^L$.
Therefore, the number of assignments associated with partition $p=(p_1,\dots,p_L)$ is at most $C(k) N^L$.
 Since for a partition to be non-singular, it must not contain $1$, \ie, $p_L\geq 2$, $L\leq\lfloor\frac{k}{2}\rfloor$ holds.
Therefore, the number of non-singular assignments is at most $C(k) N^{\lfloor\frac{k}{2}\rfloor}$.
\end{proof}

\section{Asymptotic Reusing Bias}
This section investigates the reusing bias in the asymptotic case.
We first investigate the bias in a general case in Appendix~\ref{appendix:lemmata} and then prove the propositions in Appendix~\ref{appendix:proofs-main-results}.

\subsection{Reusing Bias for Bivariate Entire Functions}\label{appendix:lemmata}
Lemma~\ref{lem:plug-in_bias_lemma} shows that the bias defined for a general bivariate entire function $\tau$~(see Definition~\ref{deff:bivariate-entire}) is $O(1/N)$.

\begin{deff}[Bivariate entire function]
\label{deff:bivariate-entire}
We refer to $\alpha\colon V_1\times V_2\to W$ as a bivariate entire function
 if $V_1$ and $V_2$ are Banach spaces
 and $\alpha$ is entire on the direct product $V_1\times V_2$.
\end{deff}

\begin{lemma}[Asymptotic reusing bias for bivariate entire functions]
    \label{lem:plug-in_bias_lemma}
    Let $\cX$ be a set.
    Let $\cP(\cX)$ denote the set of probability measures on $\cX$.
    Let $\tau(G_1, G_2)$ be a bivariate entire function on $\cP(\cX)^2$
    and define the $N$-th bias function of $\tau$ by
    \begin{align}
     \label{eq:22}b_N(G;\tau) \coloneqq \bbE_{\hG\sim G^N}[\tau(\hat{G}, \hat{G}) - \tau(\hat{G}, G)],
     \quad N\ge 1,\quad G\in \cP(\cX),
    \end{align}
    where
    $\hat{G}\coloneqq \frac1N\sum_{n=1}^N \delta_{X_n}$ denotes the empirical distribution of
    an i.i.d.~sample $\cD=\{X_n\}_{n=1}^N\sim G$.
    Then, we have
    \begin{align*}
        b_N(G;\tau)
        &=
        \frac1{2N}
     \bbE_{X\sim G}\left[2\tau^{(1,1)}_{G,G}(\delta_{X}-G,\delta_{X}-G) + \tau^{(0, 2)}_{G,G}(\delta_{X}-G,\delta_{X}-G) \right]
        +O(1/{N^2}).
    \end{align*}
\end{lemma}
\begin{proof}
    Since it is entire, $\tau$ admits the Taylor expansion everywhere,
    i.e.,
    \begin{align*}
        \tau(G_1,G_2)&=
        \sum_{k=0}^\infty
        \frac1{k!} \sum_{\ell=0}^k {k \choose \ell} \tau^{(\ell, k-\ell)}_{G,G}((G_1-G)^{\otimes \ell},(G_2-G)^{\otimes (k-\ell)})
    \end{align*}
    for all $G,G_1,G_2\in \cP(\cX)$.
    Thus, substituting $\hat{G}$ for $G_1$ and $G_2$ and taking the expectation with respect to $\hG$,
    we have
    \begin{align*}
        &\bbE[\tau(\hat{G}, \hat{G})] \\
        = &
        \sum_{k=0}^\infty
        \frac1{k!} \sum_{\ell=0}^k {k \choose \ell}
        \bbE[\tau^{(\ell, k-\ell)}_{G,G}((\hat{G}-G)^{\otimes l},(\hat{G}-G)^{\otimes (k-l)})]
        \\
        =&
        \sum_{k=0}^\infty
        \frac1{k!} \sum_{\ell=0}^k {k \choose \ell}
        \frac1{N^k}\sum_{n_1,\dots,n_k=1}^N
        \bbE\left[\tau^{(\ell, k-\ell)}_{G,G}\left(\bigotimes_{i=1}^{l}(\delta_{X_{n_i}}-G)),\bigotimes_{j=l+1}^{k}(\delta_{X_{n_j}}-G))\right)\right],
    \end{align*}
    where the first equality is owing to the entirety of $\tau$
    and the last equality follows from the multilinearity of the Fr\'echet derivatives.

 Let us calculate each of the summands using the stochastic expansion formula~(Lemma~\ref{lem:stochastic-exp-formula}) in the following.
The summand of $k=0$ is $\tau(G,G)$.
The summand of $k=1$ is calculated as,
    \begin{align*}
        &\frac1{1!} \sum_{\ell=0}^1 {1 \choose \ell}
        \frac1{N}\sum_{n_1=1}^N
     \bbE\left[\tau^{(\ell, 1-\ell)}_{G,G}\left(\bigotimes_{i=1}^{l}(\delta_{X_{n_i}}-G)),\bigotimes_{j=l+1}^{1}(\delta_{X_{n_j}}-G))\right)\right]\\
=& \frac1{N}\sum_{n_1=1}^N \bbE\left[\tau_{G,G}^{(0,1)}(\delta_{X_{n_1}} - G) + \tau_{G,G}^{(1,0)}(\delta_{X_{n_1}} - G)\right]\\
        =& 0.
    \end{align*}
    The summand of $k=2$ is calculated as
    \begin{align*}
        &\frac1{2!} \sum_{\ell=0}^2 {2 \choose \ell}
        \frac1{N^2}\sum_{n_1,n_2=1}^N
     \bbE\left[\tau^{(\ell, 2-\ell)}_{G,G}\left(\bigotimes_{i=1}^{l}(\delta_{X_{n_i}}-G)), \bigotimes_{j=l+1}^{2}(\delta_{X_{n_j}}-G))\right)\right]
        \\
        &=
        \frac1{2N^2} \sum_{\ell=0}^2 {2 \choose \ell}
        \left(\sum_{n_1=n_2}+\sum_{n_1\neq n_2}\right)\bbE\left[\tau^{(\ell, 2-\ell)}_{G,G}\left(\bigotimes_{i=1}^{l}(\delta_{X_{n_i}}-G)),\bigotimes_{j=l+1}^{2}(\delta_{X_{n_j}}-G))\right)\right] %
        \\
        &=
        \frac1{2N^2} \sum_{\ell=0}^2 {2 \choose \ell}
        \sum_{n=1}^N\bbE\left[\tau^{(\ell, 2-\ell)}_{G,G}(\delta_{X_{n}}-G, \delta_{X_n}-G)\right] %
        \\
        &=
     \frac1{2N} \sum_{\ell=0}^2 {2 \choose \ell}
     \bbE_{X\sim G}\left[\tau^{(\ell, 2-\ell)}_{G,G}(\delta_{X}-G,\delta_{X}-G)\right].
    \end{align*}
    It is also shown that
    the summands of $k\ge 3$ is $O(1/N^{\lceil \frac{k}{2} \rceil})$ by Lemma~\ref{lem:stochastic-exp-formula}.
    Summing up, we have
    \begin{align*}
        \bbE[\tau(\hat{G}, \hat{G})]
        &=
        \tau(G,G)
     +\frac1{2N} \sum_{\ell=0}^2 {2 \choose \ell}
     \bbE_{X\sim G}\left[\tau^{(\ell, 2-\ell)}_{G,G}(\delta_{X}-G,\delta_{X}-G)\right]
        +O\left(\frac1{N^2}\right).
    \end{align*}
    This procedure of asymptotic expansion is called \emph{the stochastic expansion}.
    The stochastic expansion is similarly applicable to
    the other half of the expectations in $b_N(G)$,
    \begin{align*}
        \bbE[\tau(\hat{G}, G)]
        &=
        \tau(G,G)
        +\frac1{2N}
        \bbE_{X\sim G}[\tau^{(2, 0)}_{G,G}(\delta_{X}-G, \delta_{X}-G)]
        +O\left(\frac1{N^2}\right).
    \end{align*}
    Combining these two expansions yields the desired result.
\end{proof}

The bias $b_N(G;\tau)$ in Lemma~\ref{lem:plug-in_bias_lemma} is dependent on $G$, and therefore, we cannot calculate it in practice.
By substituting $\hat{G}$ for $G$, we can construct its bootstrap estimator.
The following lemma suggests that $\tau(\hat{G},\hat{G})$ with bias correction by the bootstrap estimator $b_N(\hat{G};\tau)$ is the second-order biased estimator of $\tau(\hat{G},G)$, which is better than the plug-in estimator $\tau(\hat{G},\hat{G})$.

\begin{lemma}[Bias of a bootstrap estimator of the reusing bias]
    \label{lem:bootstrap_bias_lemma}
    Take $G$, $\hat{G}$ and $b_N(G;\tau)$ as in
    Lemma~\ref{lem:plug-in_bias_lemma}.
    Then, we have
    \begin{align*}
        \bbE[b_N(\hat{G};\tau)]
        &=
        b_N(G;\tau) + 
        O(1/{N^2}),
    \end{align*}
which implies,
 \begin{align}
  \label{eq:23}\bbE[\tau(\hat{G},\hat{G}) - b_N(\hat{G};\tau)] = \bbE[\tau(\hat{G},G)] + O(1/N^{2}).
 \end{align}
    Moreover,
    \begin{align*}
        \sqrt{\bbE\cbr{b_N(\hat{G};\tau) - b_N(G;\tau)}^2}
        &= O(1/{N^{1.5}}).
    \end{align*}
\end{lemma}
\begin{proof}
    Define the coefficient of the $O(1/N)$ term of the bias by,
    \begin{align*}
        \beta(G;\tau)\coloneqq \bbE_{X\sim G}\left[2\tau^{(1,1)}_{G,G}(\delta_{X}-G, \delta_{X}-G) + \tau^{(0, 2)}_{G,G}(\delta_{X}-G,\delta_{X}-G) \right], %
    \end{align*}
    so that
    $b_N(G;\tau)=\frac1{2N}\beta(G;\tau)+O(1/N^{2})$.
    Note that $\beta(G;\tau)$ is an entire function independent of $N$.
    Thus, the stochastic expansion of $\beta(\hat{G};\tau)$ at $G$ gives
    \begin{align*}
        \bbE[\beta(\hat{G};\tau)]
        &=
        \sum_{k=0}^\infty \frac1{k!} \bbE\left[\beta^{(k)}_G((\hat{G}-G)^{\otimes k};\tau)\right]
        \\
        &=
        \sum_{k=0}^\infty \frac1{k!} \frac1{N^k}\sum_{n_1,...,n_k=1}^N
        \bbE\left[\beta^{(k)}_G(\delta_{X_{n_1}}-G,...,\delta_{X_{n_k}}-G;\tau)\right]
        \\
        &=\beta(G;\tau) + \frac1{2N} \bbE_{X\sim G}\left[\beta^{(2)}_G((\delta_X-G)^{\otimes 2};\tau)\right] + O(1/{N^2}).
    \end{align*}
    Substituting this back to
    $\bbE[b_N(\hat{G};\tau)]=\frac1{2N}\bbE[\beta(\hat{G};\tau)]+O(1/N^2)$, we have
 \begin{align*}
  \bbE[b_N(\hat{G};\tau)]&=\frac1{2N}\left[\beta(G;\tau) + \frac1{2N} \bbE_{X\sim G}\left[\beta^{(2)}_G((\delta_X-G)^{\otimes 2};\tau)\right] + O(1/N^{2})\right] + O(1/N^{2})\\
&=\frac1{2N}\beta(G;\tau)+O(N^{-2}) = b_N(G;\tau) + O(1/N^{2}),
 \end{align*}
 which is the first desired result.

    Similarly, the stochastic expansion gives
    \begin{align*}
        &\bbE\cbr{\beta(\hat{G};\tau)-\beta(G;\tau)}^2
        \\
&= \bbE\cbrdisplay{\sum_{k=1}^\infty \frac1{k!} \beta_G^{(k)}((\hat{G}-G)^{\otimes k};\tau)}^2 \\
        &=
        \sum_{k=1}^\infty \sum_{\ell=1}^\infty
        \frac1{k!} \frac1{\ell !}
        \bbE\sbrdisplay{
            \beta^{(k)}_G((\hat{G}-G)^{\otimes k};\tau)
            \beta^{(\ell)}_G((\hat{G}-G)^{\otimes \ell };\tau)
        }
        \\
        &=
        \sum_{k=1}^\infty \sum_{\ell=1}^\infty
        \frac1{k!} \frac1{\ell !}
        \frac1{N^k} \frac1{N^\ell}
        \sum_{n_1,...,n_k=1}^N
        \sum_{n'_1,...,n'_\ell=1}^N
        \bbE\sbrdisplay{
            \beta^{(k)}_G(\delta_{X_{n_1}}-G, ..., \delta_{X_{n_k}}-G;\tau)
            \beta^{(\ell)}_G(\delta_{X_{n'_1}}-G, ..., \delta_{X_{n'_\ell}}-G;\tau)
        }
        \\
        &=\frac1{N} \bbE_{X\sim G}\sbrdisplay{\beta^{(1)}_G(\delta_X-G;\tau)\beta^{(1)}_G(\delta_X-G;\tau)} + O(1/{N^2})
        \\
        &=O(1/{N}),
    \end{align*}
    and thus
    \begin{align*}
        \sqrt{\bbE\cbr{b_N(\hat{G};\tau) - b_N(G;\tau)}^2}
        &=\sqrt{\frac{1}{4N^2}\bbE\cbr{\beta(\hat{G};\tau) - \beta(G;\tau)}^2+O\rbrdisplay{1/N^4}}
        = O\left(1/N^{1.5}\right).
    \end{align*}
\end{proof}

\subsection{Asymptotic Reusing Bias and Its Estimators}\label{appendix:proofs-main-results}
This section proves the main results presented in the main body.

\begin{proof}
 [Proof of Proposition~\ref{prop:bias}]
 By applying the composition property of entire functions~(Proposition~\ref{prop:entirety_composition}) to Lemma~\ref{lem:analyticity_of_cumulative_reward} and Assumption~\ref{assum:entire-algo},
 $\piJ(\pi(G_1), f(G_2))\colon\cP(\cS\times\bbR)^2\to\bbR$ is a bivariate entire function.
 Therefore, by applying Lemma~\ref{lem:plug-in_bias_lemma}, we obtain $b_N(G) = O(1/N)$.
\end{proof}

\begin{proof}
 [Proof of Proposition~\ref{prop:train-test-split-bias}]
Since $J(G_1,G_2)$ is entire, it can be expanded around $(G_1,G_2)=(G,G)$ and we have,
 \begin{align*}
  J(G_1,G_2)&=
  \sum_{k=0}^\infty
  \frac1{k!} \sum_{\ell=0}^k {k \choose \ell} J^{(\ell, k-\ell)}_{G,G}((G_1-G)^{\otimes \ell},(G_2-G)^{\otimes (k-\ell)}).
 \end{align*}

 Suppose we split the sample $\cD$ into $\cD_\train$ and $\cD_\test$ whose sample sizes are $N_\train$ and $N_\test$ respectively such that $\cD_\train\cup\cD_\test=\cD$, $\cD_\train\cap\cD_\test=\emptyset$, and $N_\train\colon N_\test=\lambda\colon(1-\lambda)$~($0<\lambda<1$).
 Then, we have,
 \begin{align*}
  \bbE_{\hG\sim G^N}[b_{\mathrm{split}}(\hG)] = \bbE_{\substack{\hG_\train\sim G^{N_\train}\\ \hG_\test\sim G^{N_\test}}}[J(\hG_\train, \hG_\train) - J(\hG_\train, \hG_\test)].
 \end{align*}

 Each of the terms in the right-hand side can be expanded as follows:
 \begin{align}
  \bbE_{\hG_\train\sim G^{N_\train}}J(\hG_\train, \hG_\train) =& J(G,G) + \frac1{2N_\train} \sum_{l=0}^2 {2 \choose l} \bbE_{X\sim G}\left[J_{G,G}^{(l,2-l)}(\delta_X-G, \delta_X-G)\right] + O(1/{N_\train^2}),\\
\label{eq:15}\begin{split}
  \bbE_{\substack{\hG_\train\sim G^{N_\train}\\ \hG_\test\sim G^{N_\test}}}J(\hG_\train, \hG_\test) =& J(G,G) + \frac1{2N_\train}\bbE_{X\sim G}\left[J_{G,G}^{(2,0)}(\delta_X-G, \delta_X-G)\right] \\
  &+\frac1{2N_\test} \bbE_{X\sim G} \left[J_{G,G}^{(0,2)}(\delta_X-G, \delta_X-G)\right] + O(1/N_\test^2).
\end{split}
 \end{align}
By combining the above expansions, we have,
\begin{align*}
 \bbE_{\hG\sim G^N}[b_\mathrm{split}(\hG)] =& \frac1{\lambda N} \bbE_{X\sim G}\left[J^{(1,1)}_{G,G}(\delta_X-G,\delta_X-G)\right] + \frac1{2N}\left(\frac1{\lambda}-\frac1{1-\lambda}\right)\bbE_{X\sim G}\left[J_{G,G}^{(0,2)}(\delta_X-G, \delta_X-G)\right] \\
 & + O(1/N^2) = b_N(G) + O(1/N).
\end{align*}
\end{proof}

\begin{proof}
[Proof of Proposition~\ref{prop:bootstr-bias-estim}]
As shown in the proof of Proposition~\ref{prop:bias}, $J(G_1,G_2)$ is a bivariate entire function.
By applying Lemma~\ref{lem:bootstrap_bias_lemma}, we obtain $\bbE_{\hG\sim G^N}b_N(\hG)=b_N(G)+O(1/N^2)$.
\end{proof}

\section{Reusing Bias for Optimal Policies}\label{app:reusing-bias}
This section proves that the reusing bias is non-negative for optimal policies under certain assumptions~(Proposition~\ref{prop:opt-bias}).

\begin{proof}
 [Proof of Proposition~\ref{prop:opt-bias}]
 Let $\pi(H)\coloneqq \argmax_{\pi\in\Pi} \piJ(\pi,f(H))$ for any distribution $H\in\cP(\cS^\star\times\bbR)$.
Since $\piJ(\pi(\hG),f(\hat{G})) \geq \piJ(\pi(G),f(\hat{G}))$ holds for any empirical distribution $\hG$,
taking the expectations of $\hG\sim G^N$ yields,
 \begin{align}
  \bbE_{\hG\sim G^N} \piJ(\pi(\hG),f(\hG)) \geq \bbE_{\hG\sim G^N} \piJ(\pi(G),f(\hG)) = \piJ(\pi(G),f(G)).
 \end{align}
 Since $\piJ(\pi(G),f(G)) \geq \piJ(\pi(\hG),f(G))$ holds for any $\hG$,
 taking the expectations of $\hG\sim G^N$ yields,
\begin{align}
 \piJ(\pi(G),f(G)) \geq \bbE_{\hG\sim G^N} \piJ(\pi(\hG),f(G)).
\end{align}
 By combining these, we obtain the inequality.
\end{proof}

Proposition~\ref{prop:opt-bias-is-est} shows the same statement for the importance-sampling performance estimator.
The assumpotion says that the density ratio model is consistent.

\begin{prop}[Reusing bias is optimistic for the importance-sampling performance estimator]
\label{prop:opt-bias-is-est}
 Assume,
 \begin{align*}
 \bbE_{\hG\sim G^N} \bbE_{S\sim\hG}w(S;\pi,\hG) f(S) = \bbE_{S\sim G}w(S;\pi,G) f(S),
 \end{align*}
 holds for any $\pi$ and $f$.
 Let us define $\pi^\star(\hG)=\argmax_{\pi\in\Pi}\isJ(\pi,w(\hG);\hG)$ and $\pi^\star(G)=\argmax_{\pi\in\Pi}\isJ(\pi,w(G);G)$.
 Then, the reusing bias is optimistic: $b_N(G)\geq 0$.
\end{prop}

\begin{proof}
 $\isJ(\pi,w;G) = \bbE_{S\sim G}w(S;\pi,G) f^\star(S)$.

 Since $\isJ(\pi^\star(\hG),w(\hG);\hG) \geq \isJ(\pi^\star(G),w(\hG);\hG)$ holds for any empirical distribution $\hG$,
taking the expectations of $\hG\sim G^N$ yields,
 \begin{align}
  \bbE_{\hG\sim G^N} \isJ(\pi^\star(\hG),w(\hG);\hG) \geq \bbE_{\hG\sim G^N} \isJ(\pi^\star(G),w(\hG);\hG) = \isJ(\pi^\star(G),w(G);G),
 \end{align}
 by the assumption.
 Since $\isJ(\pi^\star(G),w(G);G) \geq \isJ(\pi^\star(\hG),w(G);G)$ holds for any sample $\hG$,
 taking the expectations of $\hG\sim G^N$ yields,
\begin{align}
 \isJ(\pi^\star(G),w(G);G) \geq \bbE_{\hG\sim G^N}\isJ(\pi^\star(\hG),w(G);G).
\end{align}
 By combining these, we obtain the inequality.
\end{proof}

For the doubly-robust performance estimator, the reusing bias can be proven to be non-negative if we can assume that (i) $\bbE_{\hG\sim G^N} \hf = f(G)$, (ii) $\bbE_{\hG\sim G^N} \bbE_{S\sim\hG}w(S;\pi,\hG) f(S) = \bbE_{S\sim G}w(S;\pi,G) f(S)$ holds for any $\pi$ and $f$, and (iii) $\bbE_{\hG\sim G^N} \bbE_{S\sim \hG} w(S;\pi,\hG) f(S;\hG) = \bbE_{S\sim G} w(S;\pi,G) f(S;G)$.
However, the last assumption is less natural than the other two assumptions, and we left this for future work.

\section{Analytical Reusing-Bias Estimation}\label{sec:numer-appr-analyt}

Proposition~\ref{prop:bias} suggests an analytical expression of the $O(1/N)$-term of the reusing bias:
\begin{align*}
 b_N^{(1)}(G) = \frac1{2N}
  \bbE_{X\sim G}&\left[2J^{(1,1)}_{G,G}(\delta_X-G,\delta_X-G) + J^{(0, 2)}_{G,G}(\delta_X-G,\delta_X-G)\right].
\end{align*}
In the literature of information criteria, such an analytical expression is further embodied by substituting a concrete algorithm and/or model so as to derive a simpler expression.
Such a derivation is not possible for recent models including neural networks optimized by SGD.
 One feasible way is to compute it by approximating the Fr\'echet derivatives numerically and substituting $\hat{G}$ for $G$.

The Fr\'echet derivatives can be numerically approximated as follows:
\begin{align*}
 \tilde{J}^{(1,1)}_{G,G}(\delta_X-G,\delta_X-G)\coloneqq\frac1{4\epsilon^2}  &\left[J((1-\epsilon)G + \epsilon \delta_X, (1-\epsilon)G + \epsilon \delta_X)\right.\\
 &- J((1-\epsilon)G + \epsilon \delta_X, (1+\epsilon)G - \epsilon \delta_X)\\
 &- J((1+\epsilon)G - \epsilon \delta_X, (1-\epsilon)G + \epsilon \delta_X)\\
 & \left.+ J((1+\epsilon)G - \epsilon \delta_X, (1+\epsilon)G - \epsilon \delta_X)\right],\\
\tilde{J}_{G,G}^{(0,2)}(\delta_X-G, \delta_X-G) \coloneqq \frac1{\epsilon^2} &
 \left[J(G, (1-\epsilon)G + \epsilon \delta_X) - 2J(G,G) + J(G,(1+\epsilon)G - \epsilon \delta_X)\right]
\end{align*}
By approximating the Fr\'echet derivative numerically, we obtain,
\begin{align}
 \label{eq:5}
 \tilde{b}_N(G) = \frac1{2N} \bbE_{X\sim G} \left[2 \tilde{J}_{G,G}^{(1,1)}(\delta_X - G,\delta_X - G)
+
\tilde{J}_{G,G}^{(0,2)}(\delta_X-G, \delta_X-G)
 \right].
\end{align}
Equation~\eqref{eq:5} can be estimated by substituting $\hG$ for $G$.

The analytical bias estimation is almost equivalent to the bootstrap method in terms of statistical performance but is worse than the bootstrap method in terms of computational complexity.
In fact, while the bootstrap method requires us to train $M$ agents and $M+1$ evaluators, the analytical bias estimation requires us to train $3M$ agents and $3M$ evaluators.
Therefore, we conclude that the analytical bias estimation is not preferred to the bootstrap method in general.

\section{Train-test Split Method}\label{appendix:train-test-split}
This section investigates more on the train-test split method.
First, Proposition~\ref{prop:train-test-split} shows that the test performance estimated by the train-test split method is biased by $O(1/N)$, which is the same order as the difference between $\bbE J(\hG,G)$ and $J(G,G)$.
In contrast, the test performance estimated by bootstrap~(\eg, Equation~\eqref{eq:23}) is biased by $O(1/N^2)$, which is better than the train-test split method.

\begin{prop}
 \label{prop:train-test-split}
 Test performance estimation by train-test split also fails:
\begin{align*}
 \bbE J(\hG_\train,\hG_\test) = \bbE_{\hG\sim G^N} J(\hG, G) + O(1/N),
\end{align*}
where $\bbE_{\hG\sim G^N}J(\hG, G)= J(G,G) + O(1/N)$ holds.
\end{prop}
\begin{proof}
 Recall that we split the sample $\cD$ into $\cD_\train$ and $\cD_\test$ with sample sizes $N_\train$ and $N_\test$ such that $N_\train:N_\test=\lambda:(1-\lambda)$.
 Equation~\eqref{eq:15} suggests that,
\begin{align}
 \begin{split}
  \bbE_{\substack{\hG_\train\sim G^{N_\train}\\ \hG_\test\sim G^{N_\test}}}J(\hG_\train, \hG_\test) =& J(G,G) + \frac1{2\lambda N}\bbE_{X\sim G}\left[J_{G,G}^{(2,0)}(\delta_X-G, \delta_X-G)\right] \\
  &+\frac1{2(1-\lambda)N} \bbE_{X\sim G} \left[J_{G,G}^{(0,2)}(\delta_X-G, \delta_X-G)\right] + O(1/N^2),
\end{split}
\end{align}
 whereas the following holds:
\begin{align}
 \bbE_{\hG\sim G^N} J(\hG,G) = J(G,G) + \frac1{2N} \bbE_{X\sim G}\left[J_{G,G}^{(2,0)}(\delta_X-G, \delta_X-G)\right] + O(1/N^2).
\end{align}
 By comparing the equations above, the coefficients of $O(1/N)$ terms do not coincide, and therefore, the test performance estimated by the train-test split is biased by $O(1/N)$.
\end{proof}

The train-test split method is less biased if the test set is sufficiently large~(Proposition~\ref{prop:train-test-split-large-test}).
This proposition is not useful in practice, but it guarantees that we can estimate $J(\hG_\train,G)$ by using a sufficiently large test sample.

\begin{prop}
 \label{prop:train-test-split-large-test}
In the limit of $N_\test\to\infty$, $J(\hG_\train, \hG_\test)$ coincides with $J(\hG_\train,G)$ in expectation up to $O(1/N_\train)$-term.
\end{prop}
\begin{proof}
[Proof of Proposition~\ref{prop:train-test-split-large-test}]
 Equation~\eqref{eq:15} suggests that,
 \begin{align}
  \lim_{N_\test\to\infty}\bbE_{\substack{\hG_\train\sim G^{N_\train}\\ \hG_\test\sim G^{N_\test}}}J(\hG_\train, \hG_\test) = J(G,G) + \frac1{2N_\train} \bbE_{X\sim G}\left[J_{G,G}^{(2,0)}(\delta_X-G, \delta_X-G)\right] + O(1/N_\train^2),
 \end{align}
 which coincides with $\bbE_{\hG_\train\sim G^{N_\train}} J(\hG_\train, G)$ up to $O(1/N_\train)$ term.
\end{proof}

Let us finally discuss why the train-test split method is less accurate in our setting, whereas it is a common practice in supervised learning.
The key difference is that $J(G_1, G_2)$ is non-linear with respect to $G_2$ in our setting, while it is linear in the setting of supervised learning.
Let us re-define $J(G_1, G_2)$ as an abstract performance estimator of a data-dependent algorithm using $G_1$ evaluated by another data-dependent algorithm using $G_2$.
The evaluation of a supervised learning algorithm $f$ can be instantiated as follows:
\begin{align}
 \label{eq:13}J(G_1, G_2) = \bbE_{Z\sim G_2} \ell(f(G_1), Z),
\end{align}
where $Z=(X,Y)$ is an example and $\ell$ denotes a loss function.
The key observation is that Equation~\eqref{eq:13} is linear in $G_2$, whereas the performance estimator in Eq.~\eqref{eq:14} is in general non-linear in $G_2$.
If $J(G_1,G_2)$ is linear in $G_2$, $J^{(0,2)}_{G,G}(\delta_X-G, \delta_X-G)=0$ holds, and therefore, the train-test split estimator coincides with the true bias up to $O(1/N)$ and is biased by $O(1/N^2)$.

\section{Experimental Settings}\label{appendix:exper-sett}
In this section, we introduce the details of our experimental settings.
While we follow the environment and agent developed by \citet{pmlr-v119-gottipati20a} as much as possible, we made some modifications for consistency with our setting.
The main difference from the original environment is that we employ a finite-horizon reinforcement learning rather than an infinite-horizon RL formulation.
The environment and agent are modified accordingly.

\subsection{Environment}
The state space $\cS$ is the set of molecules.
The action space $\cA$ is the direct product of the set of reaction templates and that of reactants.
In particular, let us represent the set of reactants by their feature vectors consisting of 35 molecular descriptors used by \citet{pmlr-v119-gottipati20a}, $\cV=\{\bv_1,\dots,\bv_L\}\subset\bbR^{35}$, and the set of reaction templates by $\Omega=\{\omega_1,\dots,\omega_W\}$,
and the action space is defined as $\cA=\bbR^{35}\times\Omega$.
Given an action $(\bv,\omega_w)\in\cA$ at the current state $s_h$, the environment transits to the next state as follows.
\begin{enumerate}
 \item If the reaction template $\omega_w$ requires one reactant, the reaction template is applied to the current mol $s_h$.
\begin{enumerate}
 \item If the reaction succeeds, one of the possible products is randomly selected as the next state.
 \item If it fails, the current molecule is set as the next state.
\end{enumerate}
 \item If the reaction template $\omega_w$ requires two reactants, the next state is defined as follows.
\begin{enumerate}
 \item Assume that the current molecule is used as the first reactant.
       The set of reactants is sorted by the distance to the query vector $\bv$ in ascending order, and the second reactant is selected by the one with the smallest distance of those in the set which can be reacted with the first reactant using the reaction template $\omega_w$. As a result, a set of possible products is obtained.
 \item Assuming that the current molecules is used as the second reactant, the first reactant is selected in the same way as the previous procedure, and another set of possible products is obtained.
 \item One of these possible products is set as the next state.
\end{enumerate}
\end{enumerate}
Note that all of the reaction templates require no more than two reactants, and the above two cases cover all.

\subsection{Agent}
We employ an actor-critic architecture following the existing work~\cite{pmlr-v119-gottipati20a}.
To adapt to the finite-horizon setting, we prepare $H+1$ copies of actors and critics, and use each copy for each step.
Let $\pi_h(a\mid s;\theta_h)$ be the actor  and $Q_h(s,a;\phi_h)$ be the critic at step $h\in[H+1]$.
The learning algorithm repeatedly obtains pairs of an actor and a critic for each $h=H,H-1,\dots,0$ backwardly.
At each step, the parameters of the actor and critic are updated as follows for $t=0,1,2,\dots,T-1$:
\begin{align}
 \theta_h^{(t+1)} &\leftarrow \theta_h^{(t)} + \alpha^{(t)} \diffp{}{\theta_h} \bbE_{(s,a)\sim \bar{\cD}_h^{(t)}} \left[Q_h(s, \pi_h(s;\theta_h^{(t)});\phi_h^{(t)})\right],\\
 \phi_h^{(t+1)} &\leftarrow \phi_h^{(t)} - \beta^{(t)}\diffp{}{\phi_h} \bbE_{(s,a)\sim \bar{\cD}_h^{(t)}}\left[\left(Q_h(s,a;\phi_h^{(t)}) - r - Q_{h+1}(s^\prime, \pi_{h+1}(s^\prime;\theta_{h+1});\phi_{h+1})\right)^2\right],
\end{align}
where $\alpha^{(t)}$ and $\beta^{(t)}$ are learning rates and $\bar{\cD}_h^{(t)}$ is a mini-batch of state-action pairs at step $h$ drawn from a sample of trajectories $\bar{\cD}$.

The actor $\pi_h(a\mid s;\theta_h)$ consists of a \emph{template selector}, which receives the current state and outputs a reaction template to be aplied at the next step, followed by a \emph{reactant selector}, which receives the current state and the output of the template selector and outputs a query vector of a reactant, $\bv\in\bbR^{35}$.

The template selector consists of a Morgan fingerprint module with radius 2 and 1024 bits, followed by a fully-connected neural network with one hidden layer with 256 units, interleaved with a softplus activation except for the last layer.
The reactant selector consists of a Morgan fingerprint module with radius 2 and 1024 bits, which is then combined with the output of the template selector and is fed into a fully-connected neural network with one hidden layer with 256 units, interleaved with a softplus activation except for the last layer.

The critic network is a mapping from the current state and the outputs of the template selector and reactant selector to the estimated value of the current state and the action generated by the actor.
The current molecule is converted into a continuous vector using the Morgan fingerprint with radius 2 and 1024 bits, which are concatenated with the outputs and fed into a fully-connected neural network with one hidden layer with 256 units, interleaved with a softplus activation except for the last layer.

For the first 500 steps, we only update the parameters of the critic, fixing those of the actor, and after that, both of them are updated for another 1,500 steps.
They are optimized by AdaGrad~\cite{Duchi2011}  with initial learning rate $4\times 10^{-4}$ and batch size 64.

\subsection{Evaluators}

The reward model $f(\hG)$ is a fully-connected neural network with one hidden layer of 96 units  with softplus activations except for the last layer.
It is trained by minimizing the risk defined over $S\sim G$ by AdaGrad for $10^4$ steps with initial learning rate $10^{-3}$ and batch size 128.

The density ratio model, KuLSI, has a regularization hyperparameter $\lambda_w$. We chose it from $\{2^{-20},\dots,2^{0}\}$ by leave-one-out cross validation.

\subsection{Computational Environment}\label{appendix:comp-envir}
We implement the whole simulation in Python 3.9.0. All of the chemistry-related operations including the template-based chemical reaction is implemented by RDKit~(2021.09.3).
We used an IBM Cloud with {16$\times$2.10GHz CPU cores}, 128GB memory, and two NVIDIA Tesla P100 GPUs.

\section{Full Experimental Result}\label{sec:addit-exper-result}
\begin{figure}[t]
\centering
 \includegraphics[width=.5\hsize]{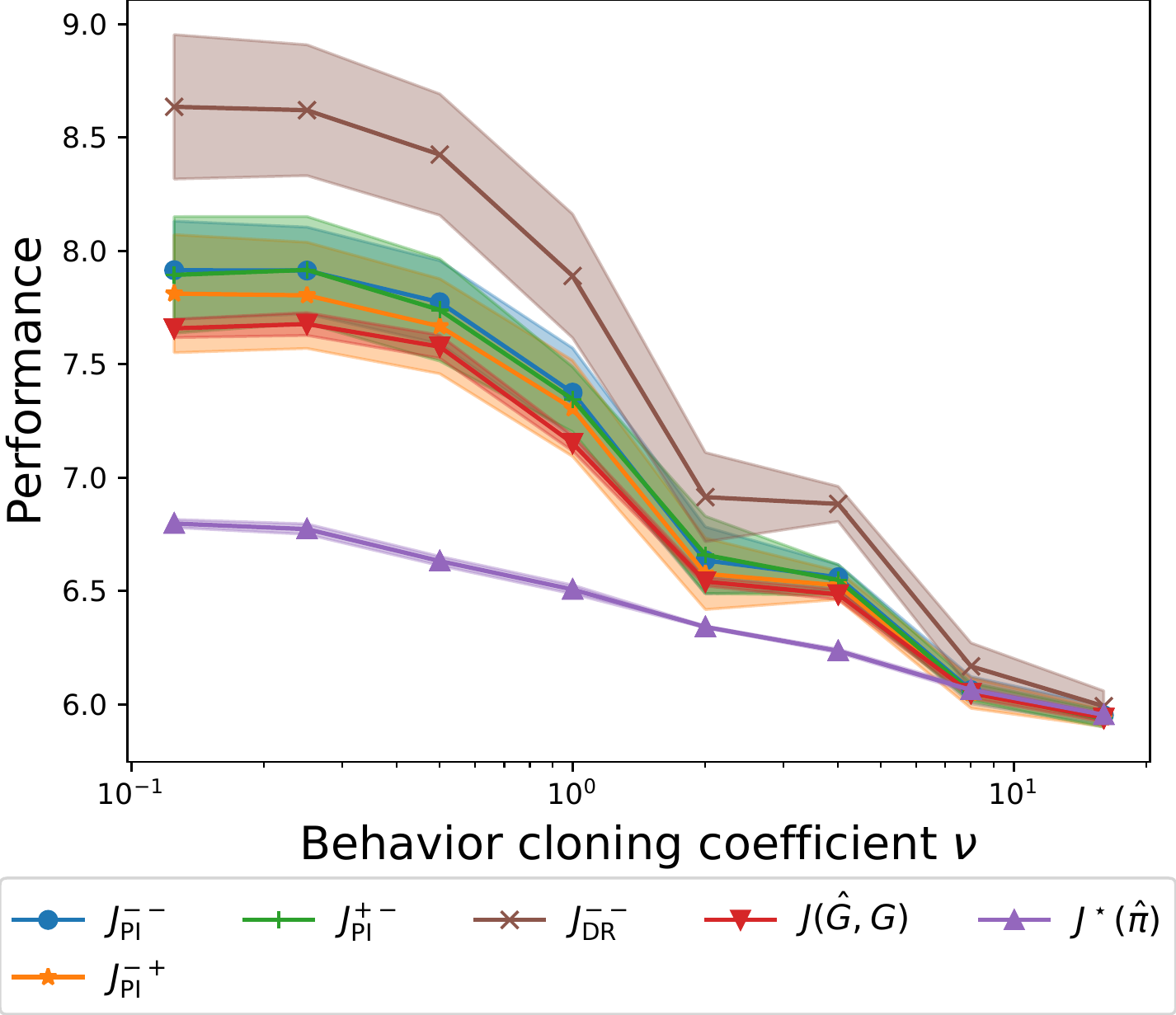}
 \caption{Experimental result with $\drJ^{--}$.}
 \label{fig:doubly-robust}
\end{figure}
We report the performance estimates by the doubly-robust performance estimator $\drJ^{--}$ in Figure~\ref{fig:doubly-robust}.
As evident from it, the scores are over-estimated primarily due to the over-estimation by the importance sampling performance estimator.

\end{document}